\documentclass[10pt,journal,cspaper,compsoc]{IEEEtran}

\ifCLASSOPTIONcompsoc
\usepackage[nocompress]{cite}
\else
\usepackage{cite}
\fi

\usepackage[nocompress]{cite}
\usepackage{graphicx}
\usepackage{amsmath}
\usepackage{amssymb}
\usepackage{multirow}
\usepackage{enumerate}
\usepackage{algorithm}
\usepackage{algorithmic}
\usepackage{slashbox}
\usepackage{array}
\usepackage{amsmath}
\usepackage[colorlinks,
			linkcolor=black,
			anchorcolor=black,
			urlcolor=black,
			citecolor=black
		   ]{hyperref}
\usepackage{bm}
\usepackage{subfig}
\usepackage{epstopdf}
\newcommand{\al}{\textit{et al. }}
\newcommand{\eg}{e.g., }

\DeclareMathOperator*{\argmin}{arg\,min}
\begin{document}

\title{Towards Automatic Construction of Diverse, High-quality Image Datasets}

\author{
	Yazhou~Yao,
	Jian~Zhang,
	Fumin~Shen,
	Li Liu,
	Fan Zhu,
	Dongxiang Zhang,
	and Heng Tao Shen
	\thanks{Y. Yao, L. Liu and F. Zhu are with the Inception Institute of Artificial Intelligence, Abu Dhabi, UAE.}
	\thanks{J. Zhang is with the Global Big Data Technologies Center, University of Technology Sydney, Australia.}
	\thanks{F. Shen, D. Zhang and H.T. Shen are with the School of Computer Science and Engineering, University of Electronic Science and Technology of China.}
    \thanks{Corresponding author: Fumin Shen (Email: fumin.shen@gmail.com).}}

\IEEEcompsoctitleabstractindextext{
\begin{abstract}
The availability of labeled image datasets has been shown critical for high-level image understanding, which continuously drives the progress of feature designing and models developing. However, constructing labeled image datasets is laborious and monotonous. To eliminate manual annotation, in this work, we propose a novel image dataset construction framework by employing multiple textual queries. We aim at collecting diverse and accurate images for given queries from the Web.
Specifically, we formulate noisy textual queries removing and noisy images filtering as a multi-view and multi-instance learning problem separately. Our proposed approach not only improves the accuracy but also enhances the diversity of the selected images. 
To verify the effectiveness of our proposed approach, we construct an image dataset with 100 categories. The experiments show significant performance gains by using the generated data of our approach on several tasks, such as image classification, cross-dataset generalization, and object detection. The proposed method also consistently outperforms existing weakly supervised and web-supervised approaches.  
\end{abstract}

\begin{IEEEkeywords}
Image dataset construction, multiple textual queries, dataset diversity
\end{IEEEkeywords}}

\maketitle
\IEEEdisplaynotcompsoctitleabstractindextext
\IEEEpeerreviewmaketitle

{\section{Introduction}}

\IEEEPARstart{A}s the computer vision community considers more visual categories and greater intra-class variations, it is clear that larger and more exhaustive datasets are needed. However, the process of constructing such datasets is laborious and monotonous. It is unlikely that the manual annotation can keep pace with the growing need for annotated datasets. Therefore, automatically constructing image datasets by using the web data has attracted broad attention \cite{hua2015prajna,schroff2011harvesting,icme2016yao,li2010optimol}.

Compared to manually labeled datasets, web images are a richer and larger resource. For arbitrary categories, the possible training data can be easily obtained from an image search engine. Unfortunately, using image search engines are limited by the poor precision of the returned images and restrictions on the total number of retrieved images. For example, Schroff \al \cite{schroff2011harvesting} reported the average precision of Google Image Search engine on 18 categories is only 39\%, and downloads are restricted to 1000 images for each query. In addition, the retrieved images usually have the overlapping problem which results in a reduced intra-class variation. In general, there are three major problems in the process of constructing image datasets by leveraging image search engine:

\textit{Scalability}. Since image search engines restrict the number of returned images for per query, Hare and Lewis \cite{hare2010automatically} proposed to adopt social network Flickr for candidate images collection while methods \cite{berg2006animals,schroff2011harvesting} addressed the problem by using a web search. In \cite{berg2006animals}, topics were discovered based on words occurring in the webpages and image clusters for each topic were formed by selecting images where the nearby text is top ranked. Then images and the associated text from these clusters were used to learn a classifier to re-rank the candidate images. These methods can obtain thousands of images for per query. However, for all of these methods, the yield is limited by the poor accuracy of the initial candidate images. 

\textit{Accuracy}. Due to the indexing errors of image search engine, even with the first few images, noise may still be included. Existing methods \cite{fergus2004visual,fergus2005learning,li2010optimol,vijayanarasimhan2008keywords} improve the accuracy by re-ranking the retrieved images. Fergus \al \cite{fergus2004visual,fergus2005learning} proposed to use visual clustering of the images over a visual vocabulary while method \cite{vijayanarasimhan2008keywords} adopted multiple instances learning to learn the visual classifiers for images re-ranking. These methods can effectively purify the error indexing images. However, for all of these methods, the yield is significantly reduced by the limited diversity of the initial candidate images which were collected with one single query.

\textit{Diversity}. Images collected with one single query tend to have a limited diversity, which is also referred as dataset bias problem \cite{yao2017exploiting,torralba2011unbiased}. To ensure the diversity of the collected images, methods \cite{vijayanarasimhan2008keywords,duan2011improving} partitioned candidate images into a set of clusters, treated each cluster as a ``bag" and the images therein as ``instances", and proposed multi-instance learning (MIL) based methods to prune noisy images. However, the yield for both of \cite{vijayanarasimhan2008keywords} and \cite{duan2011improving} is limited by the poor diversity of the initial candidate images which were obtained through one single query. To obtain lots of candidate images in a richer diversity, Divvala \al \cite{divvala2014learning} proposed to use multiple query expansions instead of a single query to collect images. However, the yield for \cite{divvala2014learning} is restricted by the iterative mechanism in the process of noise removing and images selection.

Motivated by the situation described above, we seek to automate the process of collecting images in the condition of ensuring the scalability, accuracy, and diversity. Our motivation is to leverage multiple textual queries to ensure the scalability and diversity of the collected images, and use multi-view and multi-instance learning-based methods to improve the accuracy as well as to maintain the diversity. Specifically, we first discover a set of semantically rich textual queries, from which the visual non-salient and less relevant textual queries are removed. The selected textual queries are used to retrieve sense-specific images to construct the raw image dataset. To suppress the search error and noisy textual queries (which are not filtered out) induced noisy images, we further divide the retrieved noise into three types and use different methods to filter these noise separately. To verify the effectiveness of our proposed approach, we construct an image dataset with 100 categories, which we refer to as WSID-100 (web-supervised image dataset 100). Extensive experiments on image classification, cross-dataset generalization, and object detection demonstrate the superiority of our approach. The main contributions of this work are summarized as follows: 

\begin{enumerate}

\item We propose a general image dataset construction framework that ensures the scalability, accuracy, and diversity of the image collections while with no need of manual annotation.

\item We jointly filter inter-class and intra-class noisy images in a linear programming multi-instance learning problem. Compared to existing iterative methods, our proposed approach can effectively improve the diversity while ensuring the accuracy.

\item We released our dataset on website\footnote[1]{\url{http://www.multimediauts.org/dataset/WSID-100.html}}. We hope the scalability, accuracy, and diversity of WSID-100 can help researchers further their study in the machine learning, computer vision, and other related fields.

\item We provide a benchmark platform for evaluating the performance of various algorithms in the task of pruning noise and selecting useful data. 

\end{enumerate}

This paper is an extended version of \cite{icme2016yao}, which includes about 70\% new materials. The substantial extensions include: 
treating semantic distance and visual distance as features from two different views and taking multi-view learning-based method to prune less relevant textual queries; taking MIL based method instead of iterative mechanism in the process of inter-class and intra-class noise removing; comparing the image classification ability, cross-dataset generalization ability, and object detection ability of our dataset instead of only the accuracy; comparing our dataset with both of manually labeled and web-supervised datasets instead of only manually labeled datasets; and increasing the number of categories from 10 to 100.  

The rest of the paper is organized as follows: Section 2 elaborates the related works for image dataset construction. We propose our framework and associated algorithms in Section 3. In Section 4, we compare the performance of our proposed approach with manually labeled, weakly supervised, and web-supervised baseline approaches. Section 5 concludes this paper.

\section{Related works}

Lots of works have been involved in constructing image datasets. In general, these works can be roughly divided into two types: manual based methods and learning based methods. 

\subsection{Manual Based Methods}

The traditional way to construct an image dataset is crowd based annotations (\eg ImageNet \cite{deng2009imagenet}, STL-10 \cite{coates2011analysis}, CIFAR-10 \cite{krizhevsky2009learning}, Flickr101 \cite{ballan2012combining}, YFCC100M \cite{thomee2016yfcc100m}, Caltech-101 \cite{fei2006one} and PASCAL VOC \cite{everingham2010pascal}). Most of these datasets were built by submitting a query to image search engines and aggregating retrieved images as candidate images, then cleaning candidate images by crowd annotations. The manual annotation has a high accuracy but is limited in scalability. For example, a group of students has spent several months on manually constructing the Caltech 101 \cite{griffin2007caltech} dataset. However, Caltech 101 dataset is restricted by the intraclass variation of the images (centered objects with few viewpoint changes) and the numbers of images per category (at most a few hundred). To construct the ImageNet \cite{deng2009imagenet} dataset, thousands of people have spent two years to complete. Due to the difference in knowledge, background, culture, etc., for the same category, different people often have their own tendency on choosing images, which makes the annotated dataset have a bias problem \cite{torralba2011unbiased,mm2016yao}. To ensure the diversification of the search results, ImageCLEF Photo Annotation campaign \cite{tsikrika2012building} and MediaEval Retrieving Diverse Social Images Tasks \cite{mediaeval2017} provide some standard diversification evaluation metrics.

\subsection{Learning Based Methods}

To reduce the cost of manual labeling, more and more peoples' attention has been paid to the automatic methods \cite{li2010optimol,schroff2011harvesting,hua2015prajna}. 
In \cite{li2010optimol}, Li \al took the incremental learning mechanism to collect images for the given query. It utilizes the first few retrieved images to learn classifiers, classifying images into positive or negative. When the image is classified as a positive sample, it will be used to refine the classifier. With the increase of positive images accepted by the classifier, the learned classifier will reach a robust level for this query. Schroff \al in \cite{schroff2011harvesting} proposed to adopt text information to rank retrieved images, and leverage top-ranked images to learn visual models to re-rank images once again. Hua \al \cite{hua2015prajna} leveraged clustering based method and propagation based method for pruning ``group'' and individual noisy images separately. These methods eliminate the process of manual labeling and can alleviate the scalability problem. However, for all of these methods \cite{schroff2011harvesting,li2010optimol,hua2015prajna}, the diversity of the final collected images is restricted by the limited diversity of the initial candidate images which were collected with a single query.

\begin{figure*}[t]
	\centering
	\includegraphics[width=0.95\textwidth]{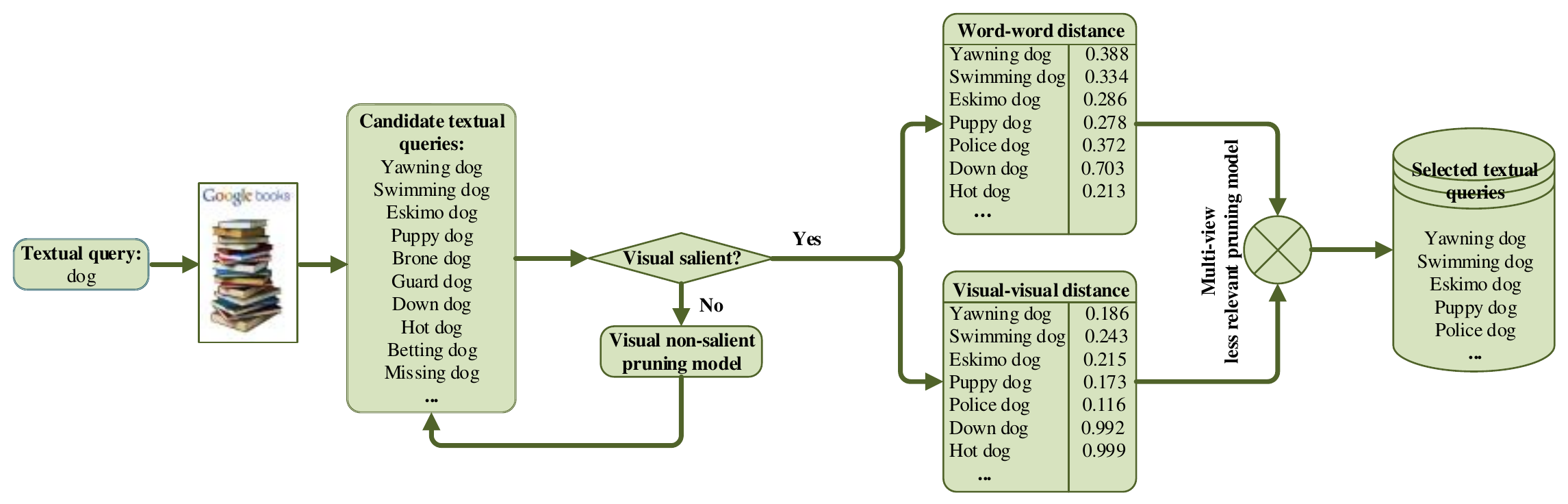}
	\caption{Illustration of the process for obtaining multiple textual queries. The input is a textual query that we would like to find multiple textual queries for. The output is a set of selected textual queries which will be used for raw image dataset construction.}
	\label{fig2}  
\end{figure*} 

\subsection{Other Related Works} 

There is a lot of work associated with the generation of multiple textual queries and noisy images removing, though their goal is not to construct an image dataset. For example, Pseudo-Relevance Feedback (PRF) \cite{valcarce2016efficient} is an automatic technique for improving the performance of a text retrieval system. Feedback information enables to improve the quality of the textual queries ranking. WordNet \cite{miller1995wordnet}, ConceptNet \cite{speer2013conceptnet} and Wikipedia are often used to obtain related synonyms for overcoming the download restriction for each query. Synonyms derived from WordNet, ConceptNet, and Wikipedia tend to be relevant to the target query and don't need to be purified \cite{li2006sentence,zhang2015memory}. 
The shortcoming is that synonyms tend to be not comprehensive enough for modifying the target query. What's worse, candidate images collected through synonyms usually have the homogenization problem, which restricts the diversity of the collected images.

To obtain diverse candidate images as well as to alleviate the homogenization problem, recent work \cite{divvala2014learning} leveraged Google Books Ngram Corpus (GBNC)\cite{lin2012syntactic} to obtain multiple textual queries for initial images collection. Compared to WordNet, ConceptNet, and Wikipedia, GBNC is much richer and general. It covers almost all related textual queries at the textual level. The disadvantage of leveraging GBNC to discover multiple textual queries is that GBNC may also bring the noise. In our work, we take GBNC to discover a set of semantically rich textual queries for modifying the target query. Then we use the word-word and visual-visual similarity to remove noisy textual queries.

A method in \cite{ballan2015data} pointed out that even for the same keyword, different search engines and social networks provide images with different styles and contents. This phenomenon may have an effect on the domain adaptation ability of the final dataset. Goodfellow \al in \cite{goodfellow2014generative} proposed a new framework for estimating generative models via an adversarial process, in which they simultaneously train two models: a generative model $ G $ that captures the data distribution, and a discriminative model $ D $ that estimates the probability that a sample came from the training data rather than $ G $. This approach can generate artificial images and opened a window for us using artificial images to do various visual tasks.	
 
\section{Framework and Methods}

We aim to propose a framework that can automatically construct image datasets in a scalable way while maintaining the accuracy and diversity of the datasets. Our proposed framework mainly consists of three major steps: candidate textual queries discovering, noisy textual queries filtering, and noisy images filtering. Specifically, as shown in Fig. \ref{fig2}, we first obtain the candidate textual queries for the target query from Google Books Ngram Corpus, from which the visually non-salient and less relevant are filtered out. As shown in Fig. \ref{fig4}, due to the indexing errors of image search engine, even we retrieve the few top images for the selected textual queries, some noise may still be included. In addition, a few noisy textual queries which are not filtered out can also bring some noise. We divide the noisy images into three groups and propose to filter out these noisy images respectively. The following subsections describe the details of our proposed framework.

\subsection{Multiple Textual Queries Discovering}

Images returned from an image search engine tend to have a relatively higher accuracy (compared to Flickr and web search), but downloads are restricted to a certain number. In addition, the accuracy of ranking-rearward images is also unsatisfactory. To overcome these restrictions, synonyms are often used to collect more images from image search engine. However, this method only works well for queries which have been defined in an existing ontology (\eg WordNet \cite{miller1995wordnet}). Apart from this, images collected by synonyms tend to have the homogenization problem \cite{torralba2011unbiased}. 

Inspired by recent work \cite{michel2011quantitative}, we can use GBNC to discover a set of semantically rich textual queries for modifying the given query. Our motivation is to leverage multiple textual queries for overcoming the download restriction of image search engine (scalability) and ensuring the greater intraclass variation of images (diversity). GBNC covers all variations of any concept the human race has ever written down in books \cite{lin2012syntactic}. Compared to WordNet and ConceptNet which only have NOUN queries, GBNC is much more general and exhaustive. Following \cite{lin2012syntactic} (see section 4.3), we specifically use the dependency gram data with parts-of-speech (POS) for refinement textual queries discovering.  
For example, given a query and its corresponding POS tag (\eg `jumping, VERB'), we find all its occurrences annotated with POS tag within the dependency gram data. Of all the gram dependencies retrieved for the given query, we choose those whose modifiers are tagged as NOUN, VERB, ADJECTIVE, and ADVERB as the candidate textual queries.
We use these semantically rich textual queries (corresponding images) to reflect the different visual distributions for the given query. The detailed candidate textual queries discovered in this step can be found on website\footnotemark[1]. 

\subsection{Noisy Textual Queries Filtering}

Multiple textual queries discovering not only brings all the useful data but also some noise (\eg ``betting dog", ``missing dog" and ``hot dog"). Using these noisy textual queries to retrieve images will have a negative effect on the accuracy. To this end, we prune these noisy textual queries before we collect candidate images for the target query. We divide the noisy textual queries into two types (visual non-salient and less relevant) and propose to filter these two types of noise separately.

\subsubsection{Visual non-salient textual queries pruning} 

From the visual consistency perspective, we want to identify visual salient and eliminate non-salient textual queries in this step (\eg ``betting dog" and ``missing dog"). The intuition is that visual salient textual queries should exhibit predictable visual distributions. Hence, we can use the image classifier-based pruning approach. 

For each textual query, we retrieve the top $N$ samples from Google Image Search Engine as positive images; then randomly split them into a training and validation set $I_i=\{I_i^t, I_i^v\}$. A pool of unrelated samples was collected as negative images. Similarly, the negative images were also split into a training and validation set $\overline I=\{\overline I^t, \overline I^v\}$. We extract 4096 dimensional deep features (based on AlexNet \cite{krizhevsky2012imagenet}) for each image and train a linear support vector machine (SVM) classifier by using $I_i^t$ and $\overline I^t$. A validation set $\{I_i^v,\overline I^v\}$ were applied to calculate the classification accuracy $S_i$. When $S_i$ takes a relatively larger value, we think textual query $i$ is visually salient. We will analyze the parameter sensitivity of $S_i$ more details in Section 4.6.

\begin{figure*}[thb]
	\centering
	\includegraphics[width=0.95\textwidth]{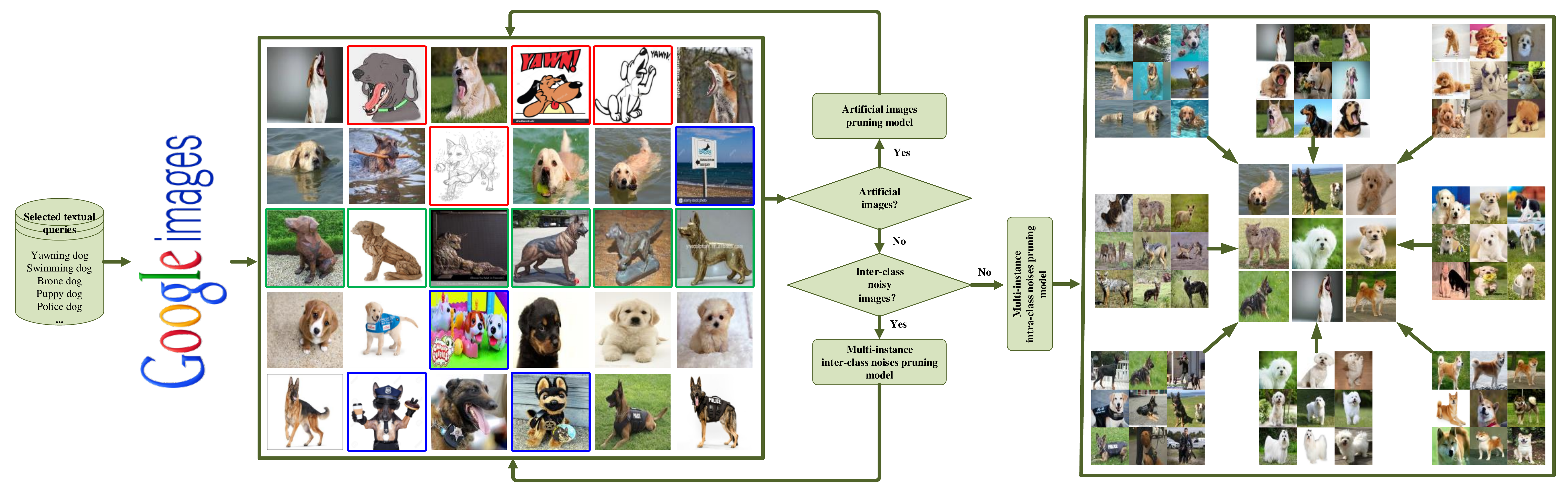}
	\caption{Illustration of the process for obtaining selected images. The input is a set of selected textual queries. Artificial images, inter-class noisy images, and intra-class noisy images are marked with red, green and blue bounding boxes separately. The output is a group of selected images in which the images corresponding to different textual queries.}
	\label{fig4}  
\end{figure*}

\subsubsection{Less relevant textual queries pruning}

Normalized Google Distance (NGD) \cite{cilibrasi2007google} extracts the semantic distance between two terms by using the Google page counts. We denote the semantic distance of all textual queries by a graph $G_{semantic}$ in which the target query is center $ y $. Another textual query $x$ has a score $S_{xy}$ corresponds to the NGD between term $x$ and $y$. Semantically relevant textual queries usually have a smaller semantic distance than less relevant (\eg ``yawning dog", ``Eskimo dog" and ``police dog" which has 0.388, 0.286 and 0.372 respectively is much smaller than ``down dog" which has 0.703). 

However, this assumption is not always true from the perspective of visual relevance. For example, ``hot dog" has a relatively smaller semantic distance 0.213, but it is not relevant to the target query ``dog". Thus, we need to identify both semantic and visual relevant textual queries for the target query.
Similar to the semantic distance, we denote the visual distance of all textual queries by graph $G_{visual}$ in which the target query is center $ y $. Another textual query $x$ has a score $V_{xy}$ corresponds to the visual distance between term $x$ and $y$. 
Similar to the previous step in Section 3.2.1, we obtain the visual distance between target query $ y $ and other textual query $x$ by the score of the center \textit{y} node classifier $ f_y $ on the \textit{x}th node retrieved images $ I_x $. The difference lies in the different test images.

By treating word-word (semantic) and visual-visual distance (visual) as features from two different views, we formulate less relevant textual queries pruning as a multi-view learning problem. Our objective is to find both semantically and visually relevant textual queries.
During training, we model each view with one classifier and jointly learn two classifiers with a regularization term that penalizes the differences between two different classifiers. 
Two views are reproducing kernel Hilbert spaces $ \mathcal{H}_{K^{(1)}} $ and $ \mathcal{H}_{K^{(2)}} $. Given $ l $ labeled data $ (x_1,y_1),...(x_l,y_l) \in \mathcal{X} \times \left \{ \pm 1 \right \} $ and $ u $ unlabeled data $ x_{l+1},...x_{l+u} \in \mathcal{X}$, we seek to find predictors $ f^{(1)*}\in \mathcal{H}_{K^{(1)}} $ and $ f^{(2)*}\in \mathcal{H}_{K^{(2)}} $ that minimize the following objective function:
\begin{equation}\label{eq5}
\begin{split}
& ( f^{(1)*},f^{(2)*}) = \argmin_{\substack{f^{(1)}\in \mathcal{H}_{K^{(1)}}\\f^{(2)}\in \mathcal{H}_{K^{(2)}}}}\mathrm{Loss}(f^{(1)},f^{(2)}) + \gamma_{1}\left \| f^{(1)} \right \|^2_{\mathcal{H}_{K^{(1)}}} \\
& + \gamma_{2}\left \| f^{(2)} \right \|^2_{\mathcal{H}_{K^{(2)}}} + \lambda \sum_{i=l+1}^{l+u}[f^{(1)}(x_i)-f^{(2)}(x_i)]^2.
\end{split}
\end{equation}
The first term is loss function and the next two are the regularization terms. The last term is called ``co-regularization" which encourages the selection of a pair predictors $ ( f^{(1)*},f^{(2)*}) $ that agree on the unlabeled data.
During testing, we make predictions by averaging the classification results from both of two views and the prediction rule is:
\begin{equation}\label{eq6}
 \mathcal{J} = \frac{1}{2}(f^{(1)}(x)+f^{(2)}(x))  
\end{equation}
Following \cite{sindhwani2005co,brefeld2006efficient}, we adopt the form of loss function as:
\begin{equation}\label{eq7}\small
\mathrm{Loss}(f^{(1)},f^{(2)}) = \frac{1}{2l}\sum_{i=1}^{l}\left ( \left [ f^{(1)}(x_i) -y_i\right ]^2 + \left [ f^{(2)}(x_i) -y_i\right ]^2 \right )  
\end{equation}
We give the solution to \eqref{eq5} in the Supplemental Material. After we obtain the models for two views, we use \eqref{eq6} to prune less relevant textual queries. 

\subsection{Noisy Images Filtering}

The selected textual queries were used to collect images from image search engine to construct the raw image dataset. Due to the indexing errors of image search engine, some noise may be included (artificial and intra-class noisy images). In addition, a few noisy textual queries which are not filtered out can also bring some noise (inter-class noisy images). As shown in Fig. \ref{fig4}, our process for filtering noisy images consists of three major steps: artificial images pruning, inter-class and intra-class noisy images pruning. 

\subsubsection{Artificial images pruning}

As we are mainly interested in constructing image datasets for natural image recognition, we would like to remove artificial images from the raw image dataset. The artificial images contain ``sketches", ``drawings", ``cartoons", ``charts", ``comics", ``graphs", ``plots" and ``maps". Since artificial images tend to have only a few colors in large areas or sharp edges in certain orientations, we choose the visual features of color and gradient histogram for separating artificial images from natural images. We train a radial basis function SVM model by using the selected visual features.
The artificial images were obtained by retrieving queries: ``sketch'', ``drawings'',``cartoons'', ``charts'', ``comics", ``graphs", ``plots" and ``maps" (250 images for each query, 2000 images in total), natural images were obtained by directly using the images in ImageNet (2000 images in total).  

After the pruning model was learned, we apply it to the entire raw image dataset to prune artificial images. 
The pruning model achieves around 94 percent classification accuracy on artificial images (using two-fold cross-validation)  and significantly reduces the number of artificial images in the raw image dataset. There is some loss of the natural images, with, on average, 6 percent removed. Although this seems to be a little high, the accuracy of the resulting dataset is greatly improved.

\subsubsection{Inter-class noisy images pruning}

Inter-class noisy images were caused by the noisy textual queries which are not filtered out. As shown in Fig. \ref{fig4} ``bronze dog" images, these noisy images tend to exist in the form of ``groups". Hence we proposed to use multi-instance learning (MIL) based method to filter these ``group" noisy images. Each selected textual query was treated as a ``bag" and the images corresponding to the textual query were treated as ``instances". We formulate inter-class noisy images pruning as a MIL problem. Our objective is to prune group noisy images (corresponding to negative ``bags").  

We denote the bags as $ B_i $, the positive and negative bags as $ B_i^+ $ and $ B_i^- $, respectively. $ \l^+ $ and $ \l^- $ denote the numbers of positive and negative bags separately. All instances belong to feature space $ \mathbb{Q} $. Bag $ B_{i} $ contains $ n_{i} $ instances $ x_{ij} $, $ j = 1,...,n_{i} $. For simplicity, we re-index instances as $ x^k $ when we line up all instances in all bags together, $ k = 1,...,n $ and $ n = \sum_{i=1}^{\l^+} n_{i}^+ + \sum_{i=1}^{\l^-} n_{i}^- $.

To characterize bags, we take the instance-based feature mapping method proposed in \cite{chen2006miles}. Specifically, we assume each bag may consist of more than one target concept and the target concept can be approximated by an instance in the bags. Under this assumption, the most-likely-cause estimator can be written as:
\begin{equation}\label{eq1}
\mathrm{Pr}(x^k|B_i) \propto s(x^k,B_i) = \max_j \exp(-\frac{\left \| x_{ij}-x^k \right \|}{\sigma^2}),
\end{equation}
where $ \sigma $ is a predefined scaling factor. $ s(x^k,B_i) $ can be explained as a similarity between bag $ B_i $ and concept $ x^k $. It is determined by the concept and the closest instance in the bag. Then the bag $ B_i $ can be embedded with coordinates 
\begin{equation}\label{eq2}
\mathbf{m}(B_i) = [s(x^1,B_i), s(x^2,B_i),...s(x^n,B_i)]^{\top}. 
\end{equation} 
Given a training set which contains $ \l^+ $ positive bags and $ \l^- $ negative bags, we apply the mapping function \eqref{eq2} and obtain the following matrix representation of all training bags:
\begin{equation}\label{eq3}
\left[
\begin{array}{ccc}
s(x^1,B_1^+) & \cdots & s(x^1,B_{\l^-}^-) \\
s(x^2,B_1^+) & \cdots & s(x^2,B_{\l^-}^-) \\
\vdots       & \ddots & \vdots            \\
s(x^n,B_1^+) & \cdots & s(x^n,B_{\l^-}^-)
\end{array}
\right].
\end{equation}
\begin{algorithm}[tb]\small
	\caption{The algorithm for learning bag classifier}
	\begin{algorithmic}[1]
		\REQUIRE ~~\\
		Positive bags $ B_i^+ $ and negative bags $ B_i^- $.\\
		\STATE \textbf{For} (each bag $ B_i = \{x_{ij}:j = 1,...,n_i\} $)\\
		\STATE \quad \textbf{for} (every instance $ x^k $) \\
		\STATE \quad \quad $ d \leftarrow  \min_{j} \left \| x_{ij}-x^k \right \| $ \\
		\STATE \quad \quad the $ k $th element of $  \mathbf{m}(B_i) $ is $ s(x^k,B^i)=e^{-\frac{d^2}{\sigma^2}} $   \\
		\STATE \quad \textbf{end}
		\STATE \textbf{End}
		\STATE Solve the linear programming  in \eqref{eq9}    \\
		\ENSURE ~~\\		
		The optimal solutions $ \mathbf{w}^* $ and $ b^* $, the bag classifier \eqref{eq11}.		
	\end{algorithmic}
	\label{alg1}
\end{algorithm}
Each column corresponds to a bag, and the $ k $th feature realizes the $ k $th row of the matrix.
Generally speaking, when $ x^k $ achieves a high similarity to some positive bags and low similarity to negative bags, we think that the feature $ s(x^k,\cdot) $ induced by $ x^k $ provides ``useful" information in separating the positive from negative bags.

Instance-based feature mapping tends to has a better generalization ability. The disadvantage is that it may require an expensive computational cost. Our solution is to construct 1-norm SVM classifiers and select important features simultaneously. The motivation is 1-norm SVM can be formulated as a linear programming (LP) problem and the computational cost will not be an issue. The 1-norm SVM is formulated as follows:
\begin{equation}\label{eq8}
\begin{aligned}
& & \min_{\mathbf{w},b,\varepsilon,\eta} \quad \lambda\sum_{k=1}^{n}\left | w_k \right | + C_1\sum_{i=1}^{\l^+}\varepsilon_i + C_2\sum_{j=1}^{\l^-} \eta_j \\
& & \mbox{s.t.} \quad (\mathbf{w}^\top \mathbf{m}_i^+ + b) + \varepsilon_i \geqslant 1, i = 1,...,\l^+, \\
& & \quad\quad -(\mathbf{w}^\top \mathbf{m}_j^- + b) + \eta_j \geqslant 1, j = 1,...,\l^-, \\
& & \quad \varepsilon_i, \eta_j \geqslant 0, i = 1,...,\l^+,j = 1,...,\l^-
\end{aligned}
\end{equation}
where $ \bm{\varepsilon} $ and $ \bm{\eta} $ are hinge losses. Choosing different parameters $ C_1 $ and $ C_2 $ will penalize on false negatives and false positives. We usually let $ C_1 = \delta $, $ C_2 = 1-\delta $ and $ 0<\delta<1 $ so that the training error is determined by a convex combination of the training errors occurred on positive bags
and on negative bags.

To solve the 1-norm SVM \eqref{eq8} with linear programming, we rewrite $ w_k = u_k - v_k  $, where $ u_k, v_k \geqslant $ 0.
Then we can formulate linear programming in variables $ \mathbf{u} $, $\mathbf{v}$, $ b $, $ \bm{\varepsilon} $ and $ \bm{\eta} $ as:
\begin{equation}\label{eq9}
\begin{aligned}
& \min_{\mathbf{u,v},b,\bm{\varepsilon,\eta}} \lambda\sum_{k=1}^{n}(u_k+v_k) + \delta\sum_{i=1}^{\l^+}\varepsilon_i + (1-\delta)\sum_{j=1}^{\l^-} \eta_j \\
& \quad \mbox{s.t.} \quad \left [ (\mathbf{u-v})^\top \mathbf{m}_i^+ + b \right ] + \varepsilon_i \geqslant 1, i = 1,...,\l^+, \\
& \quad\quad -\left [(\mathbf{u-v})^\top \mathbf{m}_j^- + b\right ] + \eta_j \geqslant 1, j = 1,...,\l^-, \\
& \quad\quad\quad \varepsilon_i, \eta_j \geqslant 0, i = 1,...,\l^+,j = 1,...,\l^- \\
& \quad\quad\quad u_k, v_k \geqslant 0, k = 1,...,n.
\end{aligned}
\end{equation}
The solutions of linear programming \eqref{eq9} equivalent to those obtained by the 1-norm SVM \eqref{eq8}. The reason is that for all $ k = 1,...,n $, any optimal solution to \eqref{eq9} has at least one of the two variables $ u_k $ and $ v_k $ equal to 0.

Suppose $ \textbf{w}^* = \mathbf{u}^* - \mathbf{v}^* $ and $ b^* $ are the solutions of \eqref{eq9}, then the influence of the $ k $th feature on the classifier can be determined by the value of $ w_k^* $. Specifically, we select features \{$ s(x^k,\cdot): k\in \phi $\} to meet the conditions: 
\begin{equation}\label{eq10}
\phi = \{k: \left|w_k^*\right| > 0\}.
\end{equation}
Finally, we obtain the classification rule of bag $ B_i $ to be positive or negative is: 
\begin{equation}\label{eq11}
y = \mathrm{sign}\left(\sum_{k\in \phi}w_k^* s(x^k,B_i)+b^*\right).
\end{equation}
The detailed process of learning the bag classifier is described in Algorithm \ref{alg1}. We apply the rule \eqref{eq11} to classify bags. When the bag is classified to be negative, the group images corresponding to the bag will be filtered out. 

\subsubsection{Intra-class noisy images pruning}

After we prune inter-class noisy images, we then only care the intra-class noise corresponding to the positive bags. The intra-class noise was induced by the indexing errors of image search engine. As shown in Fig. \ref{fig4}, this noise usually exist in the form of ``individuals".

The basic idea of pruning intra-class noise in positive bags is according to their contributions to the classification of the bag. Instances (corresponding to images) in the bags can be divided into two types: positive class and negative class. An instance is assigned to the positive class when its contribution to $\sum_{k\in \phi}w_k^* s(x^k, B_i)$ is greater than a threshold $ \theta $.
For instance $x_{ij}$ in bag $ B_i $, we define an index set $ \varphi $ as:
\begin{equation}\label{eq13}
\varphi = \left \{ j^*: j^* = \arg\max_{j}\exp\left ( -\frac{\left \| x_{ij}-x^k \right \|^2}{\sigma^2} \right ), k\in \phi  \right \}.
\end{equation}
Then the bag classification rule \eqref{eq11} only needs the instances $ x_{ij^*}, j^* \in \varphi $. Removing an instance $ x_{ij^*}, j^* \notin \varphi $ from the bag will not affect the value of $\sum_{k\in \phi}w_k^* s(x^k, B_i)$ in \eqref{eq11}. There may exist more than one instance in bag $ B_i $ maximizes $ \exp( -\frac{\left \| x_{ij}-x^k \right \|^2}{\sigma^2}  ) $ for a given $ x^k, k\in \phi $. We denote the number of maximizers for $ x^k $ by $ \nu_k $. We then rewrite the bag classification rule \eqref{eq11} in terms of the instances indexed by $ \varphi $ as:
$$ y = \mathrm{sign}\left(\sum_{j^* \in \varphi}g(x_{ij^*})+b^*\right), $$
where 
\begin{equation}\label{eq12}
g(x_{ij^*}) = \sum_{k\in \phi}\frac{w_k^* s(x^k,x_{ij^*})}{\nu _k}
\end{equation}
determines the contribution of $ x_{ij^*} $ to the classification of the bag $ B_i $. Instance $ x_{ij^*} $ belongs to the positive class if $ g(x_{ij^*}) > \theta$. Otherwise, $ x_{ij^*} $ belongs to the negative class. The choice of threshold $\theta$ is a application specific problem. In our experiments, the parameter $ \theta $ is chosen to be bag dependent as $-\frac{b^*}{\left | \varphi  \right |}$. 
The detailed process of pruning intra-class noise is described in Algorithm \ref{alg2}.
We apply the rule \eqref{eq12} to prune negative instances (corresponding to the intra-class noise).

\begin{algorithm}[tb]\small
	\caption{The algorithm for pruning intra-class noise}
	\begin{algorithmic}[1]
		\REQUIRE ~~\\
		$ \phi = \{k: \left|w_k^*\right| > 0\}  $,\\
		$ \varphi = \left \{ j^*: j^* = \arg\min_{j}\left \| x_{ij}-x^k \right \|, k\in \phi  \right \} $.\\
		\STATE Initialize $ \nu _k = 0 $ for every $ k $ in $ \phi $\\
		\STATE \textbf{For} (every $ j^* $ in $ \varphi $) \\
		\STATE \quad $ \phi_{j^*} = \{k:k\in \phi, j^* = \arg\min_{j}\left \| x_{ij}-x^k \right \| \} $ \\
		\STATE \quad \textbf{for} (every $ k $ in $ \phi_{j^*} $)   \\
		\STATE \quad \quad $ \nu_k \leftarrow \nu_k + 1  $   \\
		\STATE \quad \textbf{end}
		\STATE \textbf{End}
		\STATE \textbf{For} (every $ x_{ij^*} $ with $ j^* $ in $ \varphi $) \\
		\STATE \quad Compute $ g(x_{ij^*}) $ using \eqref{eq12} \\
		\STATE \textbf{End}
		\ENSURE ~~\\        
		All positive instances $ x_{ij^*} $ satisfying $ g(x_{ij^*}) > \theta $         
	\end{algorithmic}
	\label{alg2}
\end{algorithm}

\section{Experiments}

In this section, we construct an image dataset with 100 categories and conduct experiments on image classification, cross-dataset generalization, and object detection to verify the effectiveness of our dataset.  

\subsection{Image Dataset Construction}

We choose all the 20 categories in PASCAL VOC 2007 dataset plus 80 other categories as the target categories to construct our dataset WSID-100. The reason is existing weakly supervised and web-supervised methods were evaluated on this dataset. 

\begin{table}[t]
	\centering
	\renewcommand{\arraystretch}{1.1}
	\caption{The average accuracy (\%) comparison over 14 and 6 common categories on the VOC 2007 dataset.}
	\begin{tabular}{|p{2cm}<{\centering}|p{2cm}<{\centering}|p{2cm}<{\centering}|}
		\hline
		\multirow{2}{*}{\textbf{Method}} & \multicolumn{2}{c|}{\textbf{PASCAL VOC 2007}}  \\
		\cline{2-3}		
		&  14 categories  &   6 categories \\
		\hline	 
		STL-10 \cite{coates2011analysis}        &  -          & 39.75   \\
		CIFAR-10 \cite{krizhevsky2009learning}  &  -          & 19.04   \\
		ImageNet \cite{deng2009imagenet}        &  48.95      & 41.02   \\
		\hline
		Optimol \cite{li2010optimol}            &  42.69      & 35.97  \\
		Harvesting \cite{schroff2011harvesting} &  46.33      & 34.89  \\
		DRID-20 \cite{yao2017exploiting}        &  51.13      & 46.04  \\
		\hline
		Ours                                    &\textbf{53.88}      & \textbf{49.48}  \\
		\hline
	\end{tabular}
	\label{tab1}
\end{table}

For each category, we first discover the multiple textual queries from Google Books with POS. Then the first $ N = 100 $ images were retrieved for each discovered textual query to represent its visual distribution. In spite of the fact that noise may be contained, we treat the retrieved images as positive samples and split them into a training and validation set $I_i=\{I_i^t=75, I_i^v=25\}$. 
We gather a random pool of negative images and split them into a training and validation set $\overline I=\{\overline I^t=25, \overline I^v=25\}$. 
Through experiments, we declare a textual query $ i $ to be visual salient when the classification result $S_i \geq 0.6$. We will discuss the parameter sensitivity of $S_i$ more details in Section 4.7. We have released the discovered textual queries for 100 categories and the corresponding images (original image URL) on website\footnotemark[1].

To prune less relevant textual queries, we calculate the word-word and visual-visual distance between visual salient textual queries and target query. We label $ l_1 = 500 $ positive data and $ l_2 = 500 $ negative data. We use a total of $ l = l_1+l_2 = 1000 $ labeled and $ u = 500 $ unlabeled data to learn the multi-view prediction rule \eqref{eq6}. This labeling work only needs to be done once and the prediction rule \eqref{eq6} will be used for pruning all less relevant textual queries.

We construct the raw image dataset by using the textual queries which are not filtered out.
Specifically, we collect the top 100 images for each selected textual query. Since not enough textual queries were found for query ``potted plant", we collect the top 500 images for ``potted plant" textual query. To filter artificial images, we learn a radial basis function SVM model by using the visual feature of color and gradient histogram. Although the color and gradient histogram + SVM framework
that we use is not the prevailing state-of-the-art method for image classification, we found our method to be effective and sufficient in pruning artificial images.

\begin{figure*}[t] 
	\centering
	\subfloat[]{
		\includegraphics[width=0.24\textwidth]{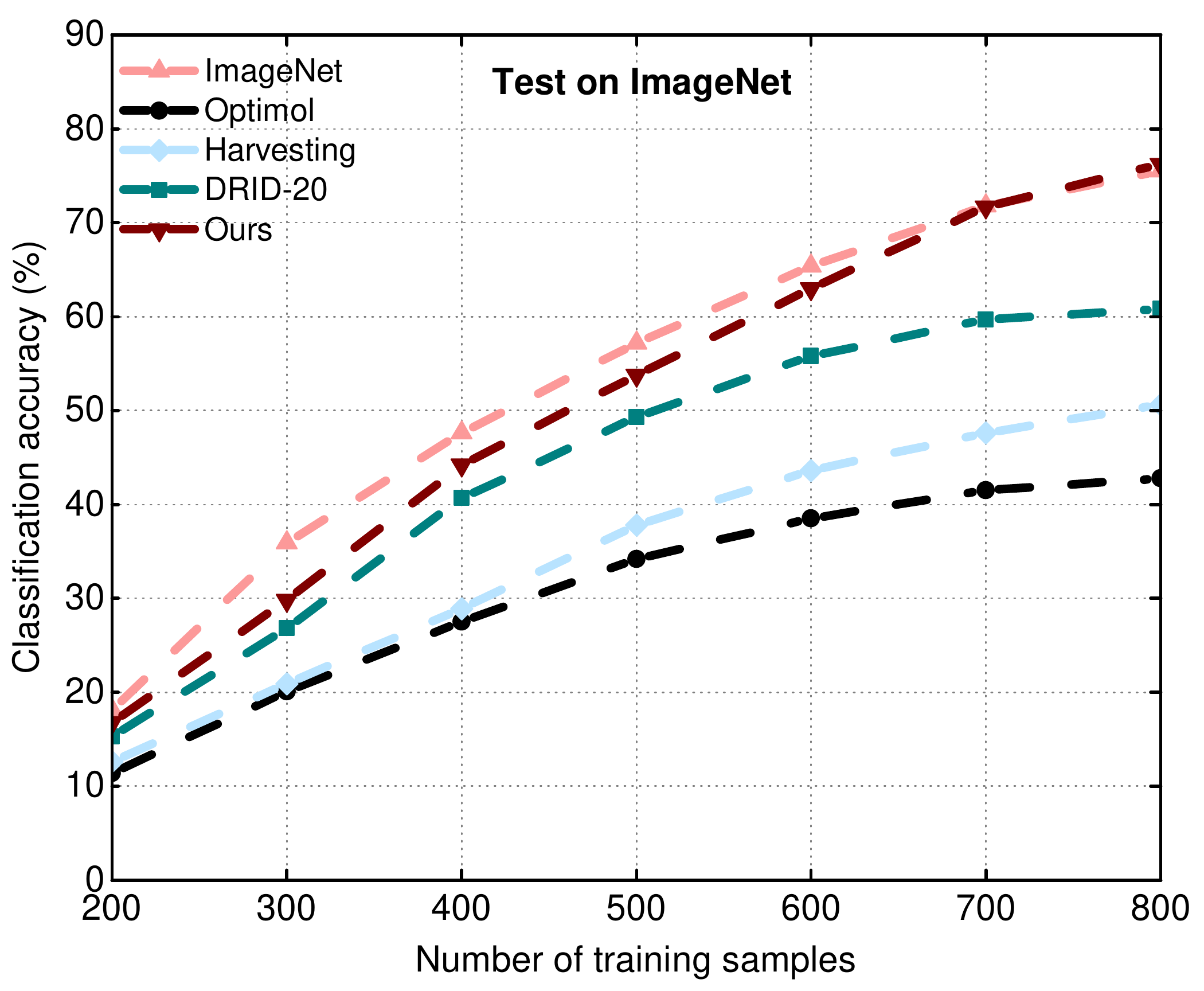}}
	\subfloat[]{
		\includegraphics[width=0.24\textwidth]{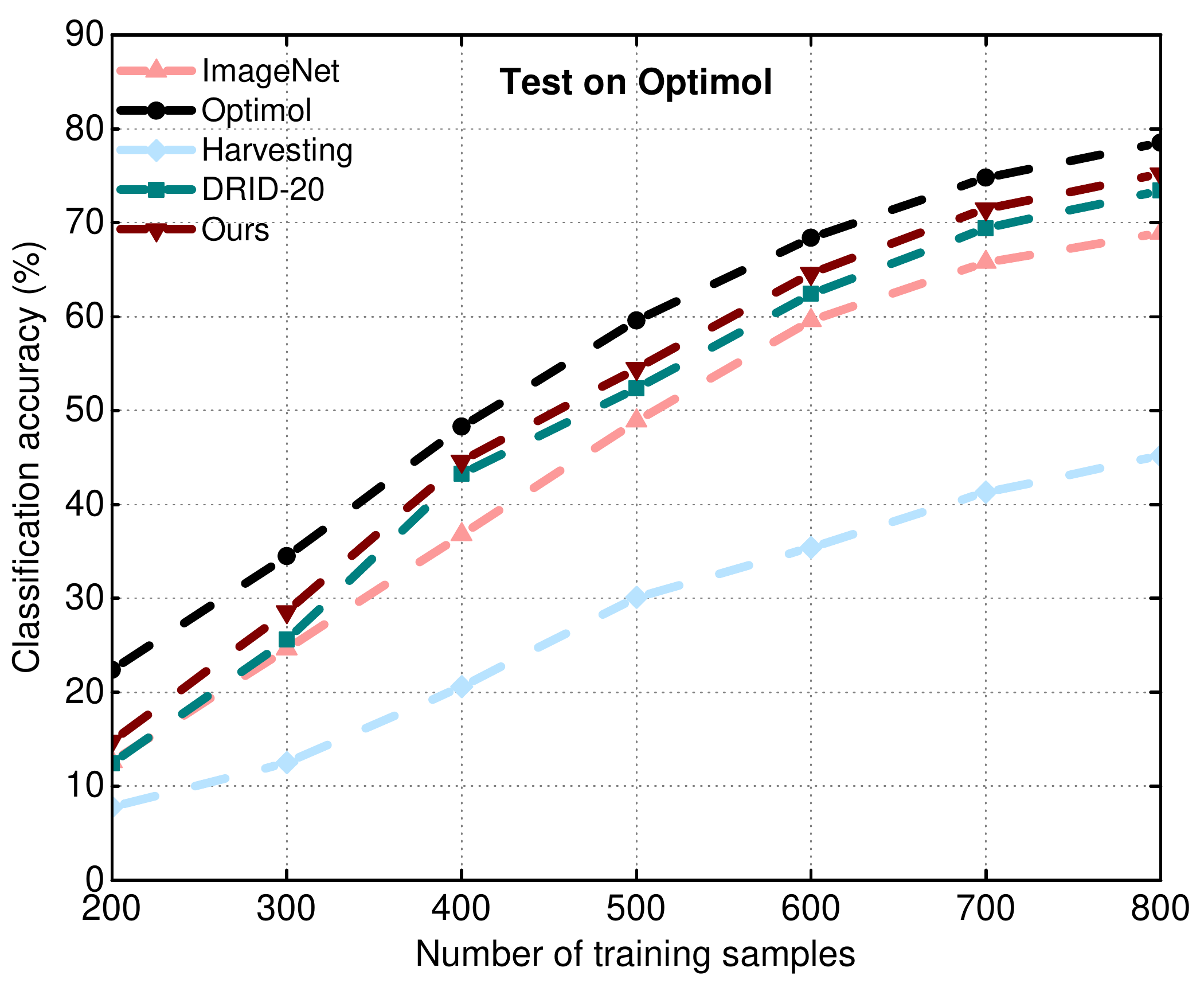}}
	\subfloat[]{
		\includegraphics[width=0.24\textwidth]{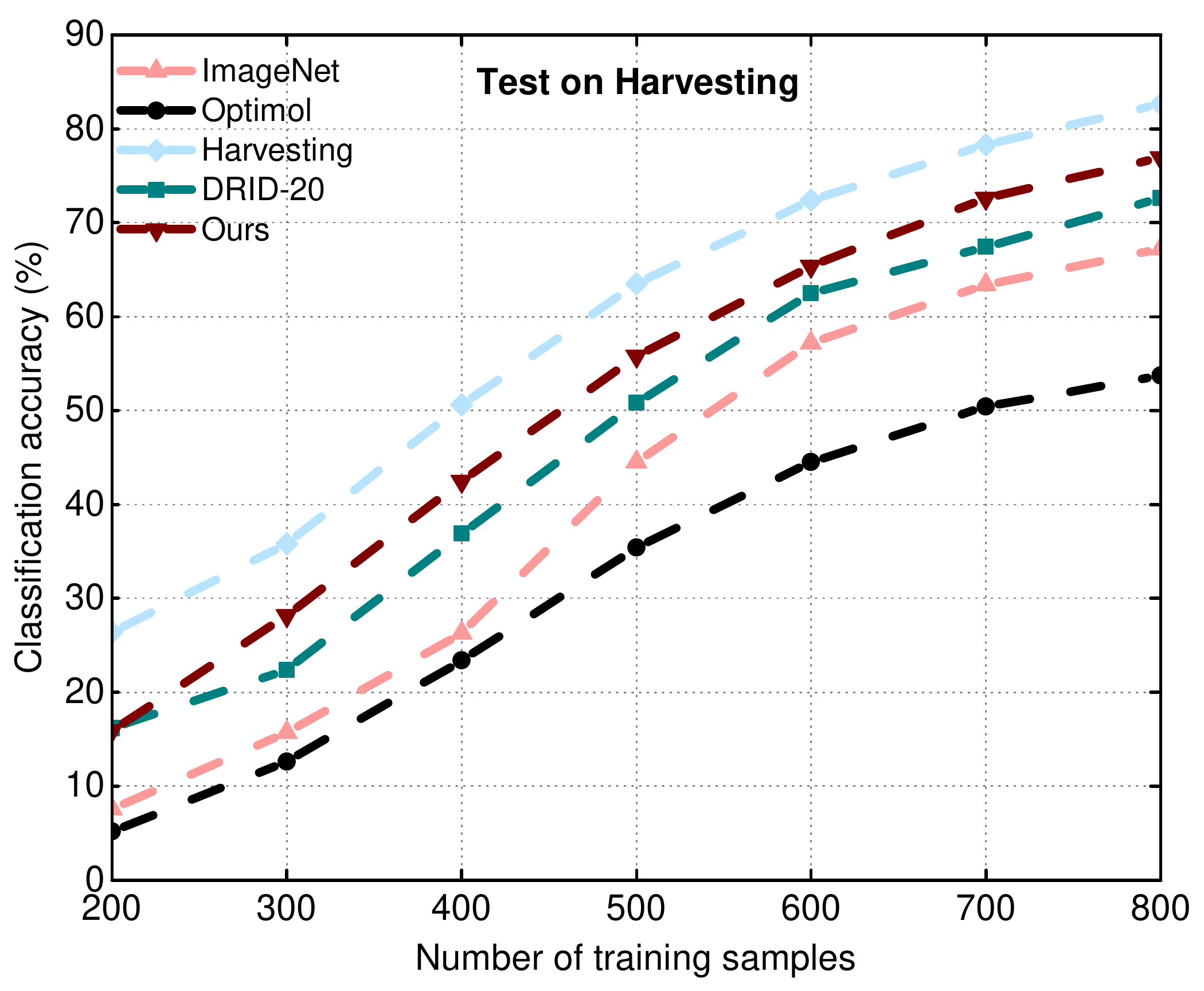}}
	\subfloat[]{
		\includegraphics[width=0.24\textwidth]{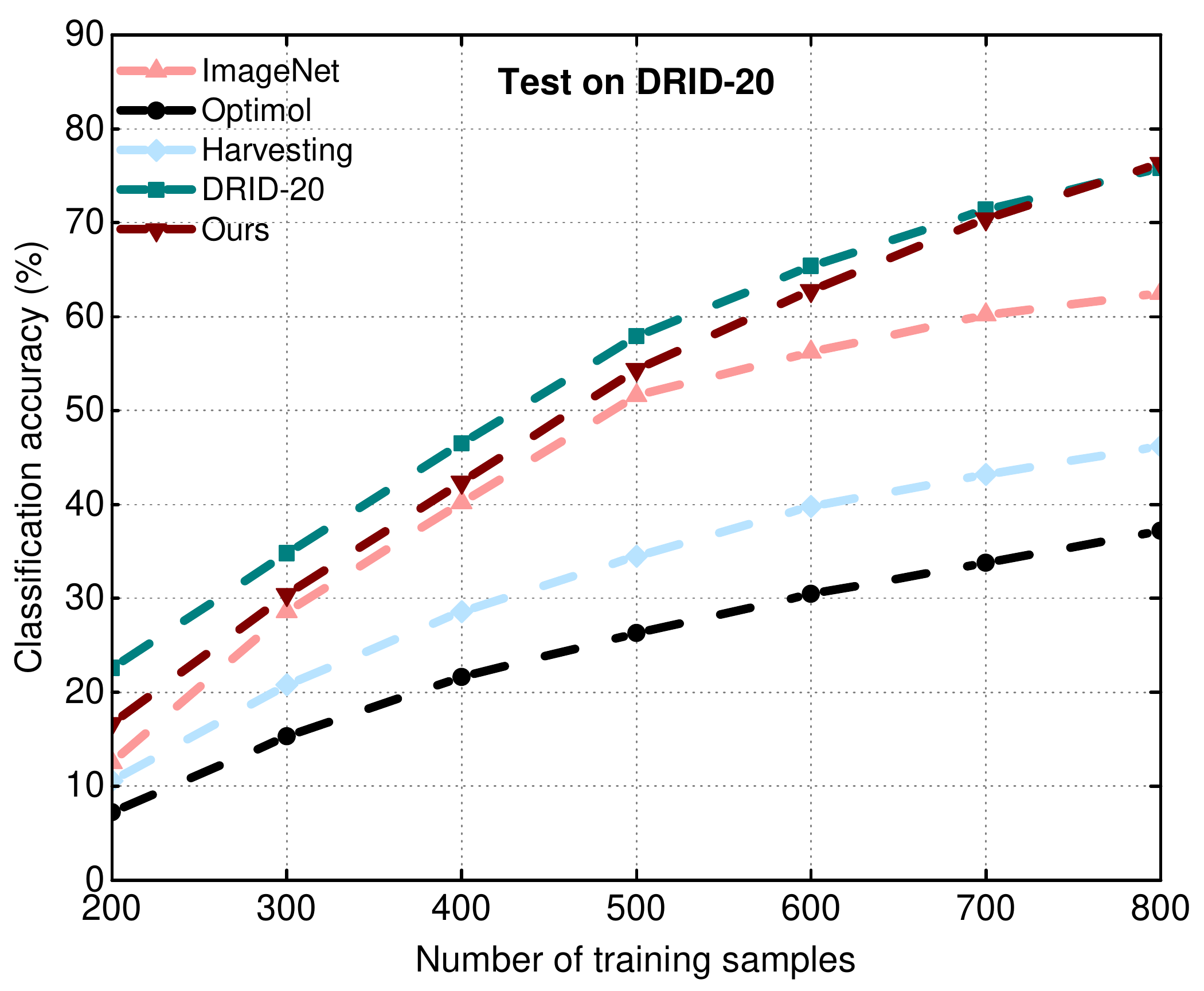}}\
	\subfloat[]{
		\includegraphics[width=0.24\textwidth]{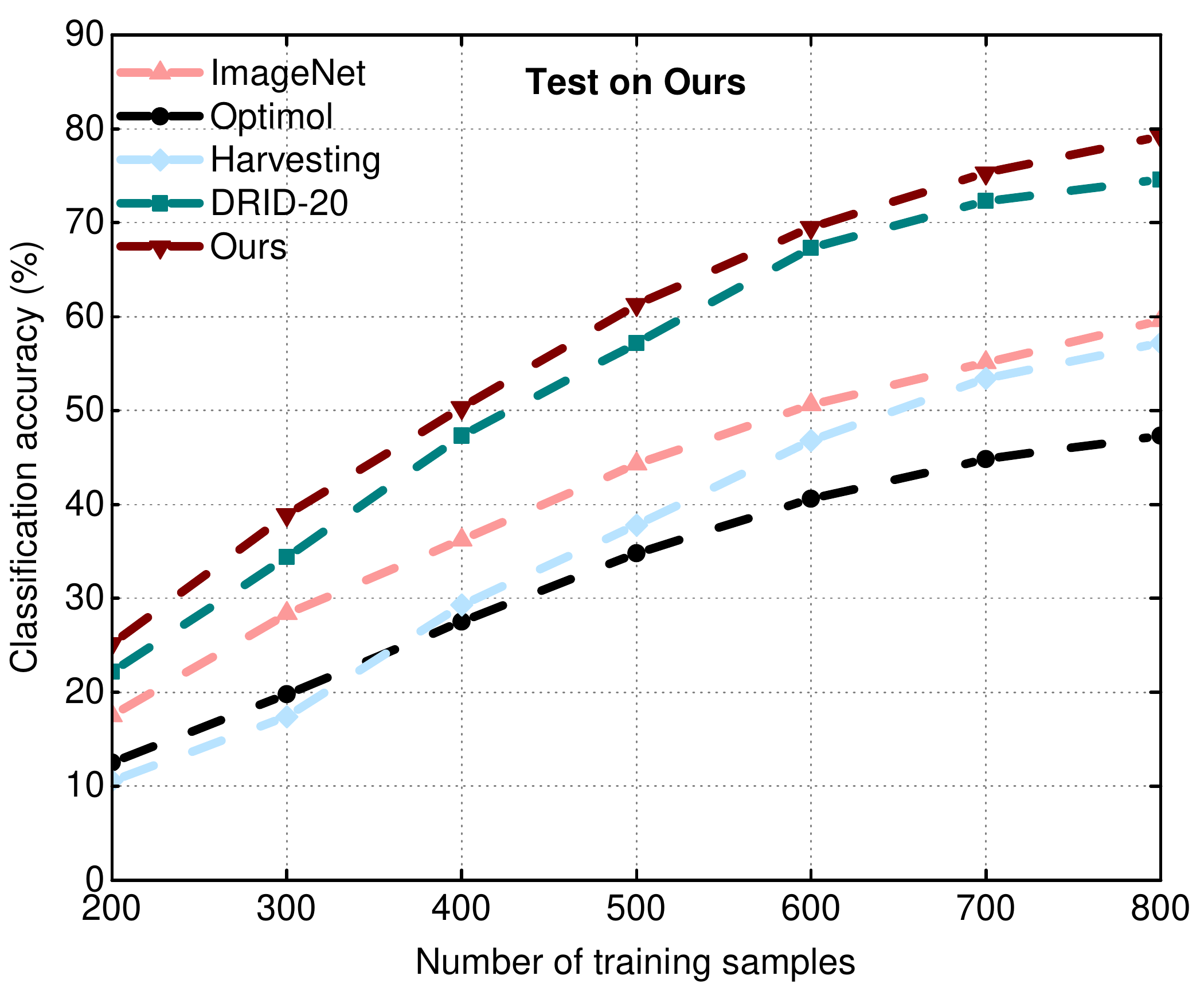}}
	\hspace{0.7cm}
	\subfloat[]{
		\includegraphics[width=0.24\textwidth]{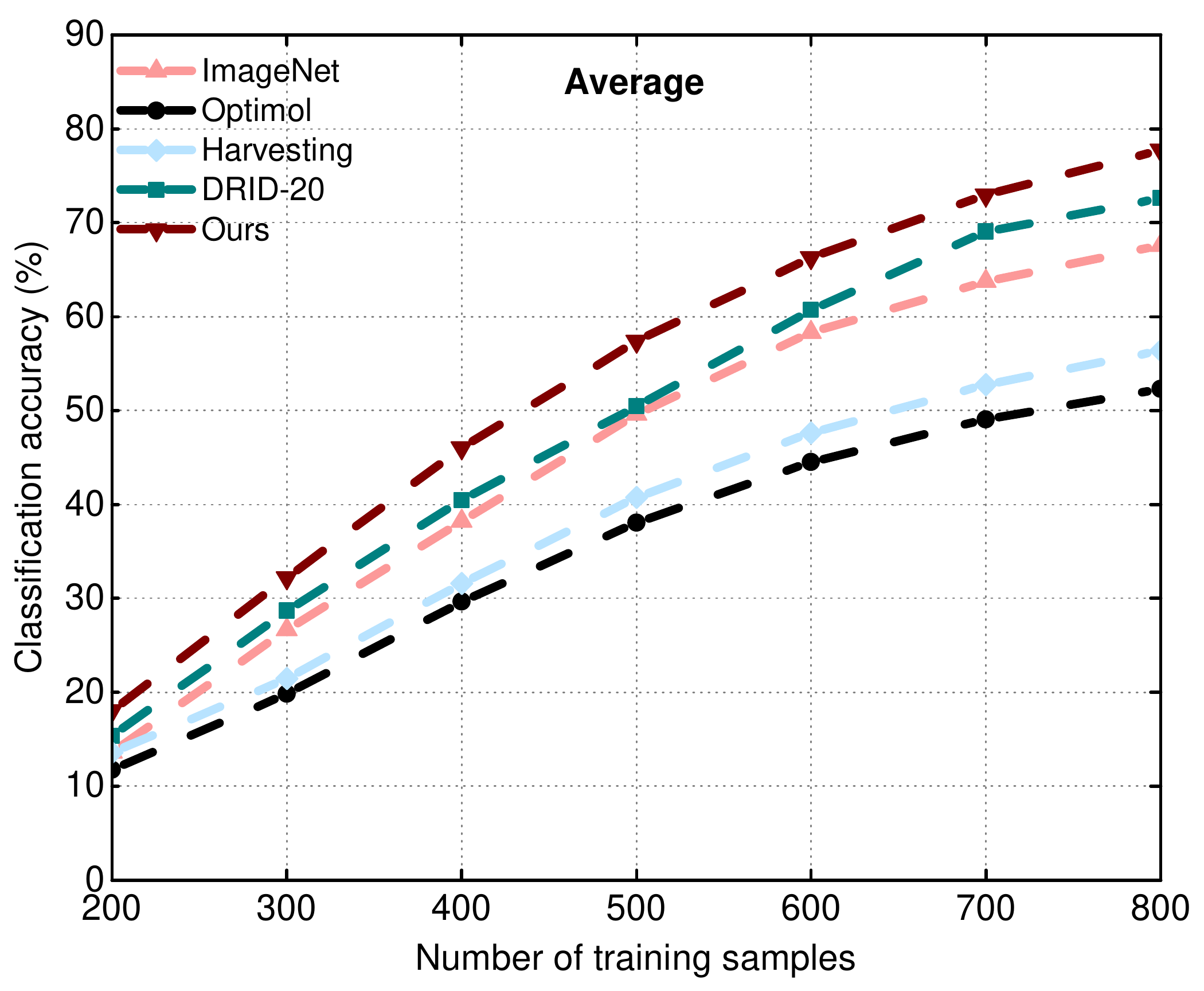}}
	\caption{The cross-dataset generalization ability of various datasets by using a varying number of training images, and tested on (a) ImageNet, (b) Optimol, (c) Harvesting, (d) DRID-20, (e) Ours, (f) Average.}
	\label{fig7}
\end{figure*}

By treating each selected textual query as a ``bag" and the images therein as ``instances", we formulate inter-class and intra-class noisy images pruning as a multi-instance learning problem. Our objective is to prune ``group" (bag-level) inter-class noisy images and ``individual" (instance-level) intra-class noisy images. To learn the bag prediction rule \eqref{eq11}, we directly use the previously labeled $ l_1 = 500 $ positive textual queries and $ l_2 = 500 $ negative textual queries corresponding images as the $ l^+ = l_1 = 500 $ positive bags and $ l^- = l_2 = 500 $ negative bags. We apply the prediction rule \eqref{eq11} to filter ``group" inter-class noisy images. 
The value of $ g(x_{ij^*}) $ in \eqref{eq12} determines the contribution of $ x_{ij^*} $ to the classification of the bag $ B_i $.
In our experiment, we choose the threshold $\theta$ as bag dependent $ \theta = -\frac{b^*}{\left | \varphi  \right |} $. That is we choose positive instance $ x_{ij^*} $ satisfying $ g(x_{ij^*}) > -\frac{b^*}{\left | \varphi  \right |} $. The value of $ b^* $ and $ \varphi $ can be obtained by solving \eqref{eq9} and \eqref{eq13}, respectively.

\subsection{Comparison of Image Classification Ability and Cross-dataset Generalization Ability}

\subsubsection{Experimental setting}

For the comparison of image classification ability, we choose PASCAL VOC 2007 \cite{everingham2010pascal} as the testing benchmark dataset. The same categories among various datasets are compared. Specifically, we randomly select 500 images for each category from various datasets as the positive training samples. 1000 unrelated images are chosen as the fixed negative samples to train SVM classification models. We test the classification ability of these models on PASCAL VOC 2007 dataset. The experiments are repeated for ten times and the average classification ability is taken as the final performance for various datasets.
The experimental results are shown in Table \ref{tab1}.

For cross-dataset generalization ability comparison, we randomly select 200 images per category from various datasets as the testing data. [200,300,400,500,600,700,800] images for each category from various datasets are sequentially chosen as the positive training samples. Similar to the comparison of image classification ability, we use the same 1000 unrelated images as the negative training samples to learn image classification models. Training and testing data for each category has no duplicates. 
Since dataset STL-10 \cite{coates2011analysis} and CIFAR-10 \cite{krizhevsky2009learning} have only 6 same categories ``airplane", ``bird", ``cat", ``dog", ``horse" and ``car/automobile" with other datasets, they won't be compared with our dataset and other datasets in this experiment. For other datasets, we compare all the 20 same categories. The average classification accuracy on all categories illustrates the cross-dataset generalization ability of one dataset on another dataset \cite{deng2009imagenet}. The experimental results are shown in Fig. \ref{fig7}.

For image classification and cross-dataset generalization ability comparison, we set the same options to learn classification models for all datasets. Specifically, we train SVM classifiers by setting the kernel as a radial basis function. The other settings use the default of LIBSVM \cite{chang2011libsvm}. For all images, we extract the 4096 dimensional deep features based on AlexNet \cite{krizhevsky2012imagenet}.   

\subsubsection{Baselines}

\textit{Manually labeled datasets}. This set of baseline methods consists of STL-10 \cite{coates2011analysis}, CIFAR-10 \cite{krizhevsky2009learning}, and ImageNet \cite{deng2009imagenet}. STL-10 contains ten categories in which per category has 500 training and 800 testing images. CIFAR-10 includes 10 categories and each category contains 6000 images. ImageNet provides an average of 1000 images to represent each category and is organized according to the WordNet hierarchy.

\begin{table*}[tb]
	\centering
	\renewcommand{\arraystretch}{1.10}\small
	\caption{Object detection results (Average Precision) (\%) on PASCAL VOC 2007 dataset (Test).}
	\begin{tabular}{cccccc|ccc}
		\hline
		\textbf{Method} & WSL \cite{siva2011weakly} & VID \cite{prest2012learning} & LEAN \cite{divvala2014learning} &  Ours & FSL \cite{felzenszwalb2010object} &  Ours-CN & Ours-FT & Fast-R \cite{renNIPS15fasterrcnn} \\
		\hline
		\textbf{Supervision}& weak     		& weak     & web           & web           & full      & web     & web       & full   \\ 
		\textbf{Detector}   & DPM     		& DPM      & DPM           & DPM           & DPM       & R-CNN   & R-CNN     & R-CNN  \\
		\hline		
		
		airplane    & 13.4     		& 17.4     & 14.0          & \textbf{17.8} & 33.2      & \textbf{30.2}     & 52.7          & 74.1\\
		
		bike        & \textbf{44.0} & -        & 36.2          & 42.4     	   & 59.0 	   & \textbf{52.6}     & 59.9          & 77.2\\
		
		bird        & 3.1      		& 9.3      & 12.5          &\textbf{ 17.7} & 10.3      & \textbf{20.7}     & 32.4          & 67.7\\
		
		boat        & 3.1      		& 9.2      & \textbf{10.3} & 9.8      	   & 15.7      & \textbf{13.3}     & 30.5          & 53.9 \\
		
		bottle      & 0.0      		& -        & 9.2           & \textbf{16.2} & 26.6      & \textbf{23.1}     & 20.9          & 51.0\\
		
		bus         & 31.2     		& -        & 35.0          & \textbf{44.6} & 52.0      & \textbf{50.6}     & 52.9          & 75.1\\
		
		car         & \textbf{43.9} & 35.7     & 35.9          & 39.7          & 53.7      & \textbf{42.4}     & 59.5          & 79.2\\
		
		cat         & 7.1      		& 9.4      & 8.4           & \textbf{11.2} & 22.5      & \textbf{22.6}     & 40.8          & 78.9\\
		
		chair       & 0.1      		& -        & \textbf{10.0} & 9.4           & 20.2      & \textbf{12.3}     & 18.6          & 50.7\\
		
		cow         & 9.3      		& 9.7      & 17.5          & \textbf{19.8} & 24.3      & \textbf{21.4}     & 43.3          & 78.0\\		
		
		table       & 9.9      		& - 		  & 6.5        & \textbf{12.3} & 26.9      & \textbf{20.0}     & 37.8          & 61.1\\
		
		dog         & 1.5      		& 3.3      & \textbf{12.9} & 12.4          & 12.6      & \textbf{21.3}     & 41.9          & 79.1\\
		
		horse       & 29.4     		& 16.2     & 30.6          & \textbf{39.5} & 56.5      & \textbf{52.4}     & 49.3          & 81.9\\
		
		motorcycle  & \textbf{38.3} & 27.3     & 27.5          & 36.3          & 48.5      & \textbf{40.9}     & 57.7          & 72.2\\
		
		person      & 4.6      		& -        & 6.0           & \textbf{8.2 } & 43.3      & \textbf{16.3}     & 38.4          & 75.9\\
		
		plant       & 0.1      		& -        & \textbf{1.5}  & 1.2           & 13.4      & \textbf{9.3}      & 22.8          & 37.2\\
		
		sheep       & 0.4      		& -        & 18.8          & \textbf{23.7} & 20.9      & \textbf{28.9}     & 45.2          & 71.4\\
		
		sofa        & 3.8      		& -        & 10.3          & \textbf{12.6} & 35.9      & \textbf{27.3}     & 37.5          & 62.5\\
		
		train       & \textbf{34.2} & 15.0     & 23.5          & 31.5          & 45.2      & \textbf{38.6}     & 48.2          & 77.4\\
		
		tv/monitor  & 0.0      		& -        & 16.4          & \textbf{20.2} & 42.1      & \textbf{28.1}     & 53.6          & 66.4\\
		\hline		
		\textbf{mAP}     	& 13.87    		& 15.25    & 17.15         & \textbf{21.32}& 33.14     & \textbf{28.62}    & 42.19         & 68.5\\				
		\hline 	    	    	    	    	    	     	       		
	\end{tabular}
	\label{tab2}
\end{table*}

\textit{Web-supervised datasets}. This set of baseline methods consists of DRID-20 \cite{yao2017exploiting}, Optimol \cite{li2010optimol} and Harvesting \cite{schroff2011harvesting}. DRID-20 contains 20 categories and each category has 1000 images. For Optimol \cite{li2010optimol}, we select all the categories in VOC 2007 as the target categories, and collect 1000 images for each category by taking the incremental learning mechanism.
For Harvesting \cite{schroff2011harvesting}, we first retrieve the possible images from the Google web search engine, and rank the retrieved images through the text information. The top-ranked images are then leveraged to learn classification models to re-rank the images once again. In total, we construct 20 same categories as VOC 2007 for Harvesting dataset. 

\subsubsection{Experimental results}

Cross-dataset generalization ability and image classification ability on third-party testing dataset measure the performance of classifiers learned from one dataset and tested on another dataset. It indicates the diversity and robustness of the dataset \cite{torralba2011unbiased,mm2016yao}.

According to the average accuracy over 6 common categories on the VOC 2007 dataset in Table \ref{tab1}, the performance of CIFAR-10 is much lower than other datasets. The explanation is that CIFAR-10 has a limited diversity and a serious dataset bias problem \cite{torralba2011unbiased}. In CIFAR-10, the objects are pure and located in the middle of the images. However, in the testing dataset and other compared datasets, these images not only consist of target objects but also plenty of other scenarios and objects. 

By observing Table \ref{tab1} and Fig. \ref{fig7}, DRID-20 has a better image classification ability and cross-dataset generalization ability than ImageNet, Optimol, and Harvesting but slightly worse than our dataset, possibly because the diversity of images in DRID-20 is relatively rich. DRID-20 was constructed by using multiple query expansions and the objects of its images have variable appearances, viewpoints, and poses. 

By observing Fig. \ref{fig7} and Table \ref{tab1}, our dataset outperforms the web-supervised and manually labeled datasets. Compared with STL-10, CIFAR-10, ImageNet, Optimol and Harvesting, our dataset constructed by multiple textual queries has a better diversity and can well adapt to third-party testing dataset.
Compared with DRID-20, our method treats textual and visual relevance as features from two different views and takes multi-view based method to leverage both textual and visual distance for pruning less relevant textual queries. Our method can be more effective in pruning textual queries and then obtain a more accurate dataset. At the same time, we convert the inter-class and intra-class noise pruning into solving a linear programming problem, not only improves the accuracy but also the efficiency.

\subsection{Comparison of Object Detection Ability}

Our goal is to demonstrate that the automatically generated dataset is meaningful. For this, we will train two kinds of detectors: 1) First, we will train the Deformable Part Models (DPM) \cite{felzenszwalb2010object} detectors. 2) We will train the Faster R-CNN \cite{renNIPS15fasterrcnn} detectors. Since recently state-of-the-art web-supervised and weakly supervised methods have been evaluated on PASCAL VOC 2007 dataset, we also test the object detection ability of our DPM and Faster R-CNN detectors on this dataset.

\subsubsection{Experimental setting for DPM detectors} 

We firstly remove images which have extreme aspect ratios ($>$ 2.5 or $<$ 0.4) and resize images to a maximum of 500 pixels.
Then we train a separate DPM for each selected textual query to constrain the visual variance. 
Specifically, we initialize our bounding box with a sub-image in the process of latent re-clustering to avoid getting stuck to the image boundary. Following \cite{felzenszwalb2010object}, we take the aspect-ratio heuristic method to initialize our components. Some components across different textual queries detectors share visual similar patterns (\eg ``police dog" and ``guard dog" ). We take the method proposed in \cite{divvala2014learning} to merge visual similar and select representative components. 
After we obtain the representative components, we leverage the approach proposed in \cite{felzenszwalb2010object} to augment and subsequently generate the final detector.

Our final detection model is a multi-component model. Given a test image, there could be several valid detections by our model, \eg an image of horse-eating grass would not only have the horse profile detection but also the ``horse head" detection. As the VOC criterion demands a single unique detection box for each test instance that has 50\% overlap, all the other valid detections are declared as false-positives either due to poor localization or multiple detections.

\subsubsection{Experimental setting for Faster R-CNN detectors}

For Faster R-CNN detectors training, we will do two experiments: 1) First, we will train Faster R-CNN detectors \cite{renNIPS15fasterrcnn} using the collected image dataset and evaluate them on the VOC 2007 dataset. Note in this case, we will not use any VOC training images. 2) We use the approach similar to R-CNN \cite{Girshick2014} where we will fine-tune our learned CNN using VOC data. 

Since the collected image dataset has no bounding box to localize objects, we now describe our strategy for localizing objects without manual annotation. We use the full images in collected image dataset as seed bounding boxes. This is mainly based on retrieved images from Google Image Search Engine have the bias toward a centered object and a clean background. For each seed, we train an Exemplar-LDA \cite{hariharan2012discriminative} detector. This Exemplar-LDA detector is then performed on the remaining images to find its top $ k $ nearest neighbors. For efficiency, instead of checking all possible windows on each image, we use EdgeBox \cite{zitnick2014edge} to propose candidate ones, which also reduces background noise. We set $ k $=10 in our experiments. We then use a publicly available variant of agglomerative clustering \cite{chen2014enriching} where the nearest neighbor sets are merged iteratively from the bottom up to form the final subcategories based on Exemplar-LDA similarity scores and density estimation.
Finally, we train a Faster R-CNN detector for each category based on all the clustered bounding boxes. We set the batch size as 256, and start with a learning rate of 0.01. We reduce the learning rate by a factor of 10 after every 100K iterations and we stop training at 500K iterations. 

We test our trained CNN model for object detection on the VOC 2007 dataset. Specifically, we follow the pipeline in R-CNN \cite{Girshick2014} and perform two sets of experiments. First, we directly test the detection ability of the learned CNN without fine-tuning on VOC data. Second, we fine-tune the CNN model by back-propagating the error end-to-end using VOC train/validation set. We performed the fine-tuning procedure 200K iteration with a step size of 20K. We named the two detectors as Ours-CN (with only collected web data) and Ours-FT (fine-tuned with VOC data). 

\subsubsection{Baselines}

\textit{Weakly supervised methods}. This set of baselines consists of \cite{siva2011weakly} and \cite{prest2012learning}. Method \cite{siva2011weakly} leverages image-level labels for training and initializes from objectness. Method \cite{prest2012learning} takes manually labeled videos without bounding box for training and presents the results in 10 categories. 
 
\textit{Web-supervised methods}. The web-supervised method \cite{divvala2014learning} leverages web information as a supervisor to train a mixture DPM detector. 

\textit{Fully supervised method}. The fully supervised methods \cite{felzenszwalb2010object,renNIPS15fasterrcnn} are a possible upper bound for weakly supervised and web-supervised methods.

\subsubsection{Experimental results}

Table \ref{tab2} presents the object detection results of our proposed approaches and other state-of-the-art methods on the VOC 2007 test set. From Table \ref{tab2}, we have the following observations: 

Compared with method \cite{siva2011weakly} and \cite{prest2012learning} which leverages weak supervision and \cite{felzenszwalb2010object} which requires full supervision, our method and \cite{divvala2014learning} don't need to label the training data. Nonetheless, our method and \cite{divvala2014learning} achieve better detection results than previously best weakly supervised methods \cite{siva2011weakly} and \cite{prest2012learning}. Compared to method \cite{divvala2014learning} which also leverages multiple textual queries for images collection and web supervision, our method achieves the best results in most cases. Possibly because we take different methods to filter noisy textual queries and images. Method \cite{divvala2014learning} takes iterative approaches during the process of noisy textual queries and images removing while our method leverages a multi-view based method for noisy textual queries removing and multi-instance learning-based method for noisy images removing. 
Our method can obtain a better diversity of the selected images in the condition of ensuring the accuracy.
Our method discovers much richer as well as more useful linkages to visual descriptions for the target category.

For the same training data (case Ours and Ours-CN), the performance of R-CNN model is much better than the traditional DPM model. The average increase was 34\%. One possible explanation is that CNN can extract more powerful feature representations, so that even with the same data, we can get a much better performance.

Using the VOC data to fine tune the detectors trained with web data can effectively improve the performance of the detector. However, there are still some gaps in the performance of detectors generated with full-supervised data. The explanation is that our training data comes from the web and may contain noise. In addition, the web training data has no manually labeled bounding boxes may also affect the performance of the detector.

\begin{figure*} [t]
	\centering
	\subfloat[]{
		\includegraphics[width=0.242\textwidth]{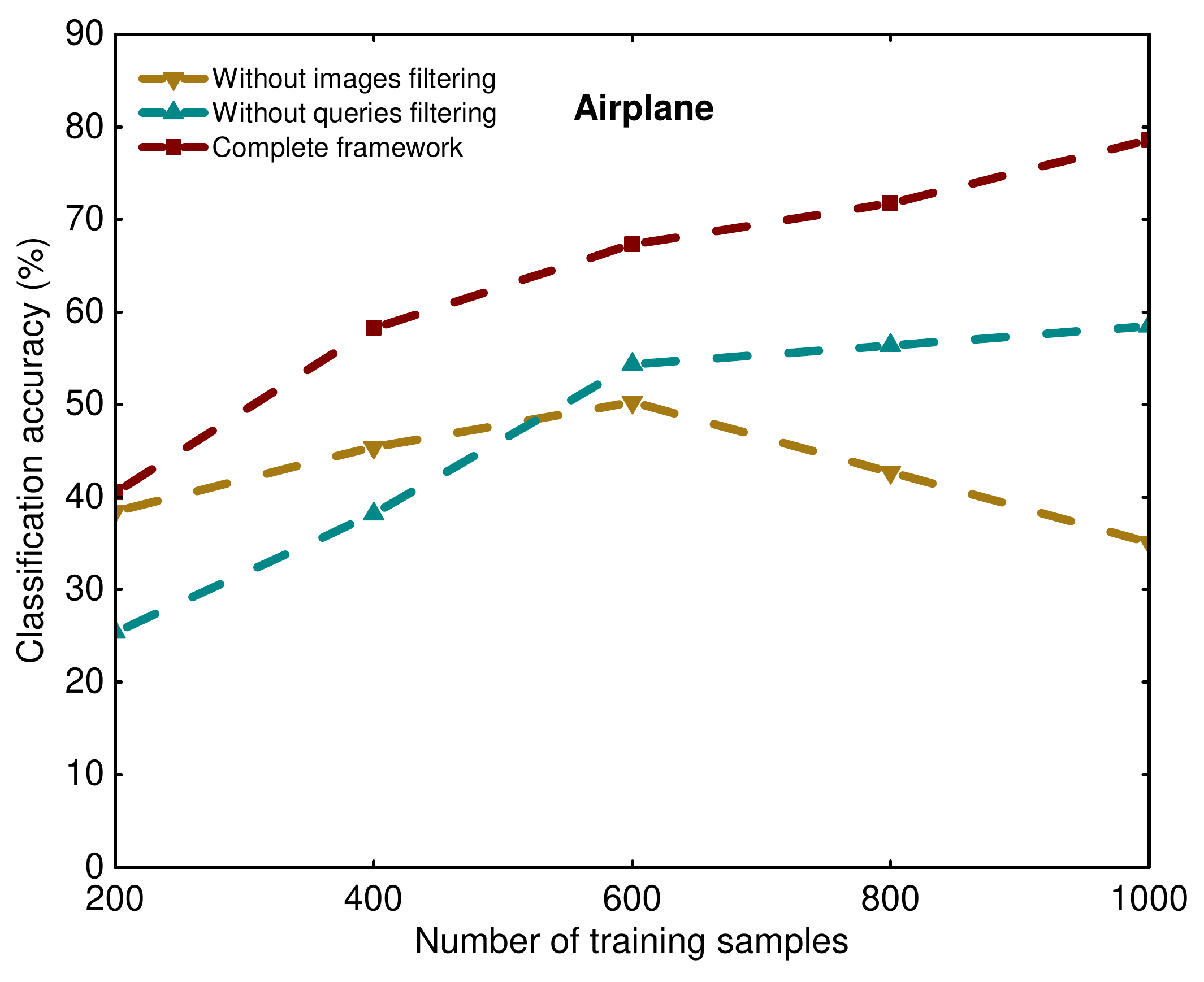}}
	\subfloat[]{		
		\includegraphics[width=0.242\textwidth]{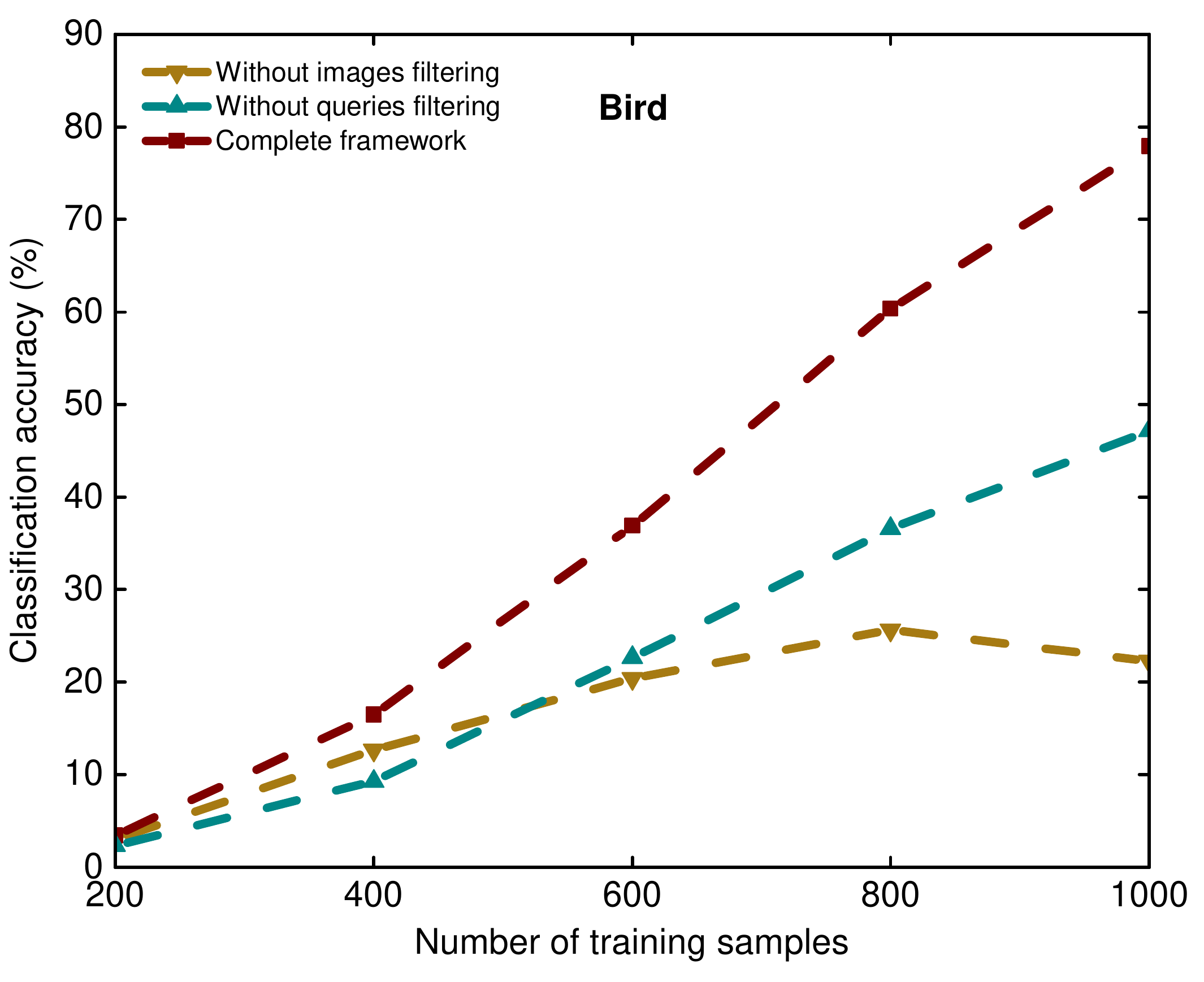}}
	\subfloat[]{
		\includegraphics[width=0.242\textwidth]{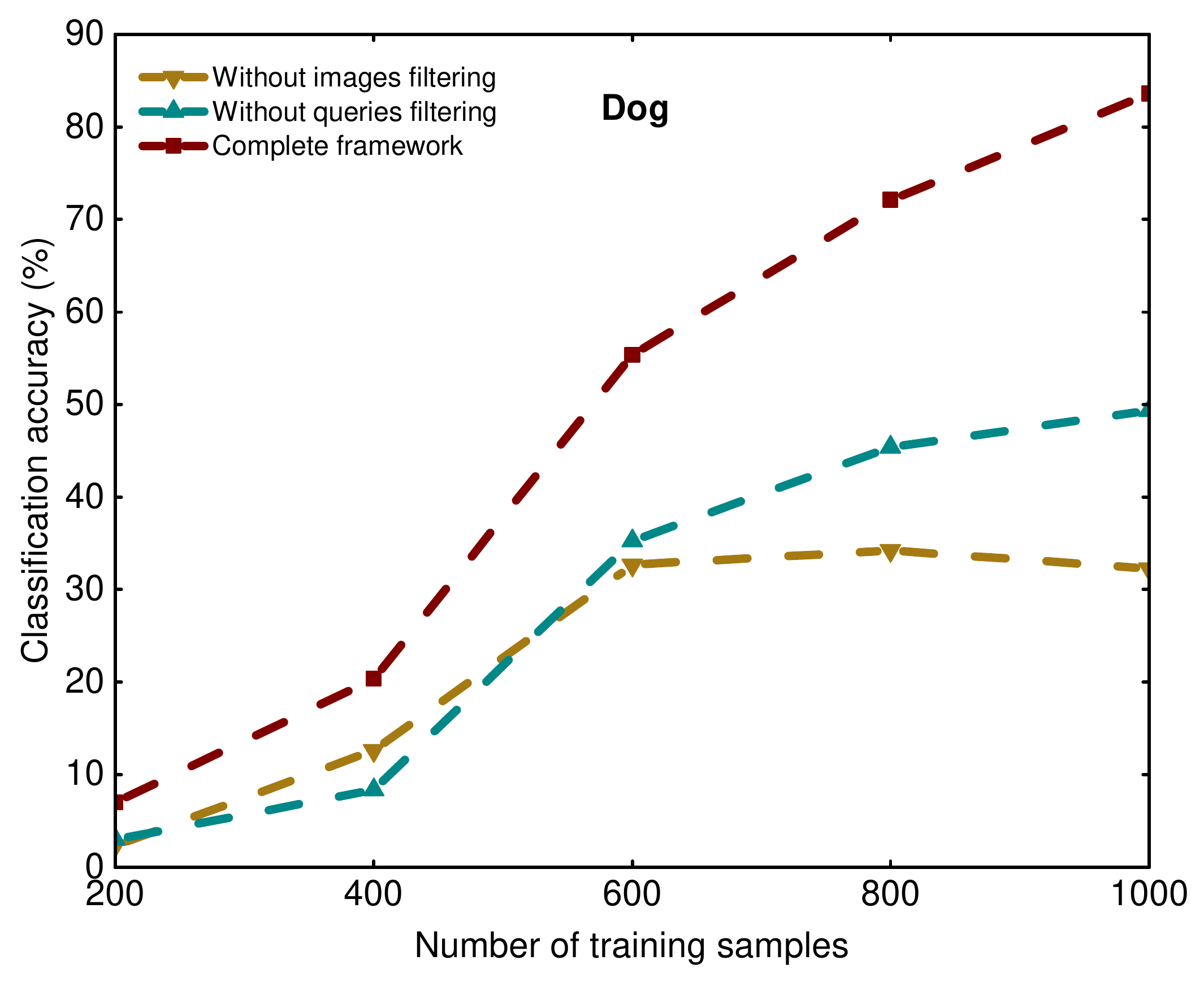}}
	\subfloat[]{		
		\includegraphics[width=0.242\textwidth]{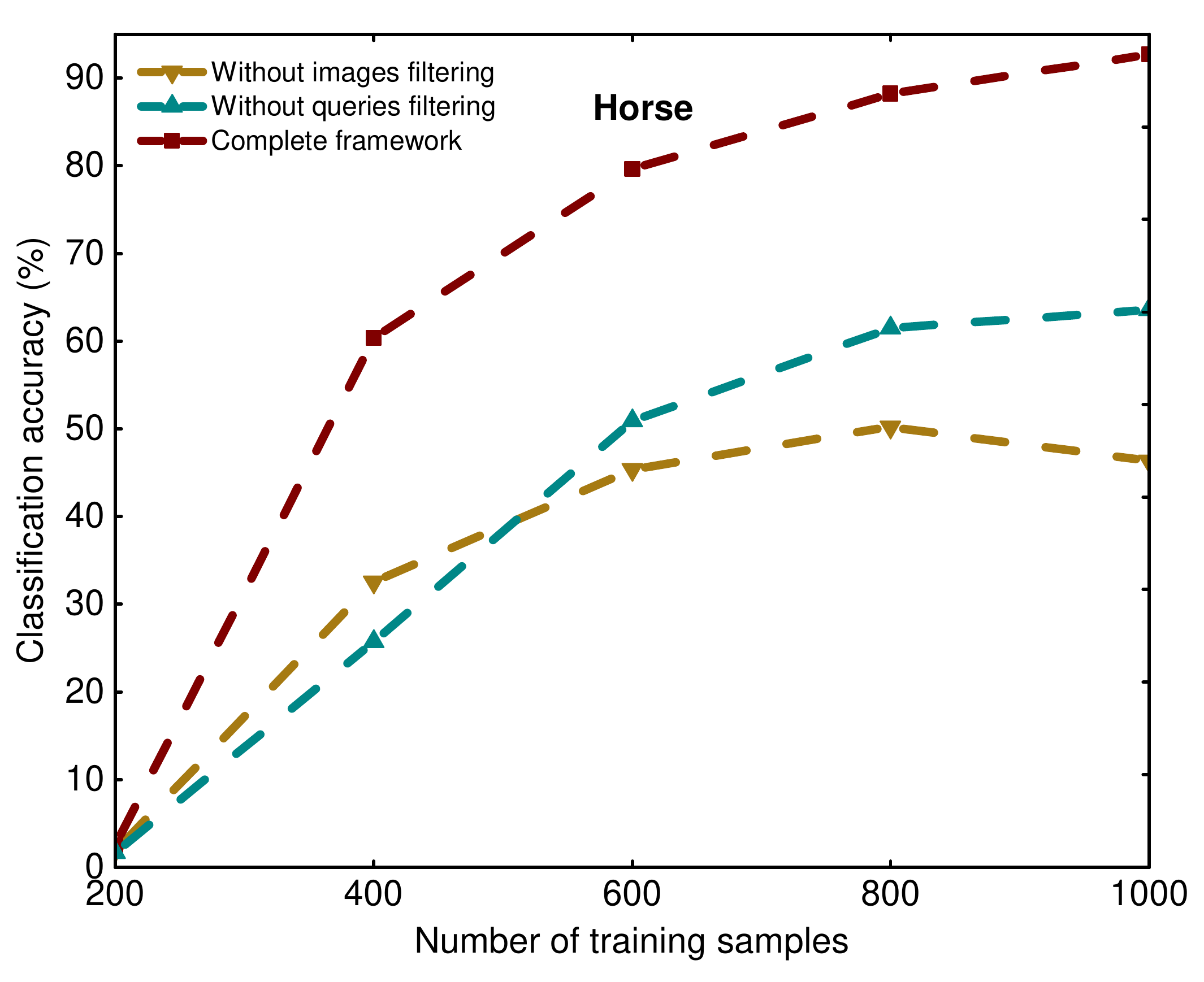}}
	\caption{Image classification ability of ``without images filtering", ``without queries filtering", and ``complete framework" on PASCAL VOC 2007 dataset: (a) ``airplane", (b) ``bird", (c) ``dog" and (d) ``horse".}
	\label{fig8}
\end{figure*}

\subsection{Different Steps Analysis}

Our proposed framework involves three major steps: multiple textual queries discovering, noisy textual queries filtering and noisy images filtering. To quantify the contribution of various steps to the final result, we construct two new frameworks. 

\begin{table}[t]
	\centering
	\renewcommand{\arraystretch}{1.10}\small
	\caption{Object detection results (A.P.) (\%) of five categories on VOC 2007 dataset (Test) with DPM detector.}
	\begin{tabular}{p{1.95cm}<{\centering}|p{0.8cm}<{\centering}p{0.65cm}<{\centering}p{0.57cm}<{\centering}p{0.57cm}<{\centering}p{0.7cm}<{\centering}p{0.57cm}<{\centering}}
		\hline
		\multirow{2}{*}{\textbf{Dataset}} & \multicolumn{6}{c}{\textbf{Category}}  \\
		\cline{2-7}		
		&  airplane   &   bird   &   dog    &    bus    &    horse   &  \textbf{mAP}       \\
		\hline
		WSID-Flickr      &  6.3        &   3.6    &   6.5    &    22.8   &    18.4    &     11.52           \\
		WSID-Bing        &  15.4       &   12.5   &   9.7    &    37.4   &    35.7    &     22.14           \\
		WSID-Google      &  17.8       &   17.7   &   12.4   &    44.6   &    39.5    &     26.40           \\
		\hline  
	\end{tabular}
	\label{tab2-2-1}
\end{table}

One is based on \textit{multiple textual queries discovering} and \textit{noisy textual queries filtering} (which we refer to ``without images filtering"). The other is based on \textit{multiple textual queries discovering} and \textit{noisy images filtering} (which we refer to ``without queries filtering"). For framework ``without images filtering", we directly retrieve the top images from the image search engine for selected textual queries to train image classifiers (without noisy images filtering). For framework ``without queries filtering", we directly retrieve the top images from the image search engine for all candidate textual queries (without noisy textual queries filtering). We apply the noisy images filtering procedure to select useful images and train image classifiers.

The image classification ability among framework ``without images filtering", ``without queries filtering" and ours are compared. 
Specifically, category ``airplane", ``bird", ``dog" and ``horse" are selected as target categories to compare the image classification ability.
We sequentially collect [200,400,600,800,1000] images per category as the positive data and leverage 1000 unrelated images as the negative data to train image classification models. We evaluate the image classification ability on the VOC 2007 dataset. The experimental results are presented in Fig. \ref{fig8}. From Fig. \ref{fig8}, we have the following observations: 

Framework ``without images filtering" usually performs better than ``without queries filtering" when the training image for each category is less than 600. One possible explanation is that the first few retrieved images tend to present a relatively high accuracy. When the number of training images is below 600, the final selected noisy images caused by noisy textual queries are more severe than the image search engine. 
As the increase of numbers per category, the retrieved images contain more and more noise.
In this condition, the noise induced by 
the indexing errors of image search engine presents a much worse influence than those caused by the noisy textual queries. 

\begin{table}[t]
	\centering
	\renewcommand{\arraystretch}{1.10}\small
	\caption{Object detection results (A.P.) (\%) of five categories on VOC 2007 dataset (Test) with R-CNN detector.}
	\begin{tabular}{p{1.95cm}<{\centering}|p{0.8cm}<{\centering}p{0.65cm}<{\centering}p{0.57cm}<{\centering}p{0.57cm}<{\centering}p{0.7cm}<{\centering}p{0.57cm}<{\centering}}
		\hline
		\multirow{2}{*}{\textbf{Dataset}} & \multicolumn{6}{c}{\textbf{Category}}  \\
		\cline{2-7}		
		&  airplane   &   bird   &   dog    &    bus    &    horse   &  \textbf{mAP}       \\
		\hline
		WSID-Flickr      &  18.5       &   11.6   &   13.5   &   35.2    &    25.6    &     22.88           \\
		WSID-Bing        &  27.2       &   18.3   &   17.4   &   47.3    &    46.7    &     31.38           \\
		WSID-Google      &  30.2       &   20.7   &   21.3   &   50.6    &    52.4    &     35.04           \\
		\hline  
	\end{tabular}
	\label{tab2-3}
\end{table}
Our proposed framework outperforms other two frameworks. The reason can be explained that our approach leverages a combination of noisy textual queries and images removing, can be effective in filtering the noise caused by both noisy textual queries and the indexing errors of image search engine.

\subsection{Different Domains Analysis}

To analyze the impact of different domains on building datasets, we constructed three image datasets with five categories for each dataset. The three datasets have the same category and size, but the image sources are from the Google Image Search Engine, the Bing Image Search Engine, and Flickr respectively. We named these three datasets as WSID-Google, WSID-Bing, and WSID-Flickr. Specifically, category ``airplane", ``bird", ``dog", ``bus",  and ``horse" are selected as target categories to compare the object detection ability. It should be noted that the data in the WSID comes from the Google Image Search Engine, so we only need to construct the Bing and Flickr data source image datasets. 

\begin{figure*}[tbp]
	\centering
	\includegraphics[width=0.99\textwidth]{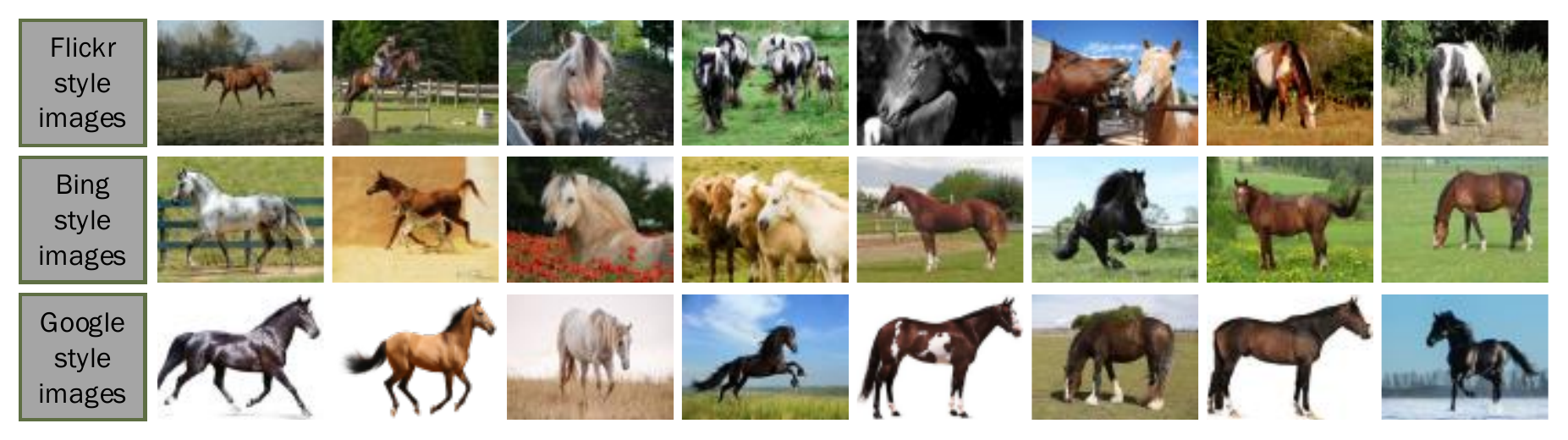}
	\caption{Different styles of images from different data sources. \eg the first line of Flickr-style images, the second line of Bing-style images, and the third line of Google-style images.}
	\label{fi1}  
\end{figure*}

To demonstrate the characteristics of different web sources, we constructed the datasets WSID-Bing and WSID-Flickr according to the process of building WSID-Google. Since the images in datasets WSID-Bing and WSID-Flickr also have no bounding boxes, and to fairly compare with dataset WSID-Google, we take the same strategy mentioned above for localizing objects in the datasets WSID-Bing and WSID-Flickr. Similarly, we also chose to train the DPM and R-CNN detector. We test it in the VOC 2007 dataset and the experimental results are shown in Table \ref{tab2-2-1} and \ref{tab2-3} respectively. From Table \ref{tab2-2-1} and \ref{tab2-3}, we have the following observations:

\begin{table}[tb]
	\centering
	\renewcommand{\arraystretch}{1.1}
	\caption{The average recall and precision for ten categories corresponding to different $S_i$ }
	\begin{tabular}{c|ccccc}
		\hline
		$S_i$    & 0.8     & 0.7     & 0.6    & 0.5   & 0.4     \\
		\hline		
		Recall   & 35.6\% & 72.3\% & 97.4\%& 98.7\% & 100\%  \\		
		Precision& 87.2\% & 78.8\% & 71.2\%& 52.7\%& 46.4\%  \\		                           
		\hline		
	\end{tabular}
	\label{tab3}
\end{table}

The performance of WSID-Flickr is much lower than WSID-Bing and WSID-Google by using both DPM and R-CNN detector. One possible explanation is that the Flickr's image data comes from people's daily life, and the background of the images is more complicated, making it difficult to accurately locate the bounding boxes of the target objects without manual labeling. Therefore, the object detection ability of WSID-Flickr is much lower than WSID-Bing and WSID-Google. 

The performance of WSID-Google is a little better than WSID-Bing by using both DPM and R-CNN detector. As shown in Fig. \ref{fi1}, the explanation may be that Google's bias toward images with a single centered object and a clean background. This allows us to obtain the bounding boxes of the target objects easily and accurately.

\subsection{Parameter Sensitivity Analysis}

There are lots of parameters in the process of our experiments, we mainly analyze two parameters $ S_i $ and $ \delta $  in our proposed framework ($ C_1 = \delta $, $ C_2 = 1-\delta $ and $ 0<\delta<1 $).
To analyze parameter $ S_i $ and $ \delta $, we choose 10 categories and manually label 50 textual queries for each category. For each textual query, we retrieve the top 100 images from image search engine to represent the visual distribution. The value of $ S_i $ is selected from the set of \{0.4, 0.5, 0.6, 0.7, 0.8\} by applying the 3-fold cross-validation method. Table \ref{tab3} demonstrates the average recall and precision for 10 categories corresponding to different $ S_i $. Finally, we choose the value of $ S_i $ to be 0.6. The reason is we want to get a relatively higher recall while ensuring an acceptable precision. 

For the parameter $ \delta $, the value is selected from \{$ 10^{-3}, 10^{-2},..., 10^{1} $\}. We also use the 3-fold cross-validation to select the value of $ \delta $. Table \ref{tab4} shows the average accuracy of inter-class noisy images filtering. 
By observing Table \ref{tab4}, we found our method is robust to the parameter $ \delta $ when it is varied in a certain range.

\subsection{Potential Applications}

Due to the cost of manual labeling is too expensive, crawling data from the Internet and using the web data (without manual annotation) to train models for various computer vision tasks have attracted broad attention. However, due to the complex of the Internet, the crawled data tend to have noise. Removing noise and choosing high-quality instances for training often plays a key role in the quality of the last trained model.

\begin{table}[tb]
	\centering
	\renewcommand{\arraystretch}{1.65}
	\caption{The average accuracy of inter-class noise filtering for ten categories corresponding to different $ \delta $ }
	\begin{tabular}{c|cccccc}
		\hline
		$\delta$    & $ 10^{-3} $     & $ 10^{-2} $     & $ 10^{-1} $    & $ 10^{0} $   & $ 10^{1} $     \\
		\hline		
		Accuracy   & 96.2\% & 97.5\% & 96.6\%& 98.2\% & 97.6\%  \\		
		\hline		
	\end{tabular}
	\label{tab4}
\end{table}

We give an example of how to use our dataset to evaluate the performance of various algorithms in the task of pruning noise. The specific steps are as follows: (1) obtaining the raw image data for 100 categories from our website; (2) performing algorithms to prune noise and select useful data from the raw image data; (3) running cross-dataset generalization experiments on the selected data and our publicly released dataset WSID-100. Algorithms which have a better cross-dataset generalization ability tend to have a better ability in the task of pruning noise and selecting high-quality data.

Weakly supervised learning algorithms (\eg MIL) is gaining interest since it allows to leverage loosely labeled data. Therefore, it has been used in diverse application fields such as computer vision and document classification. Most of the existing large-scale datasets in computer vision (\eg ImageNet) are manually labeled and fewer are weakly labeled. Thus, it is difficult to evaluate the performance and robustness of various weakly supervised algorithms. To this end, we would like to supply a benchmark dataset for evaluating the performance and robustness of various weakly supervised algorithms. The training data are from dataset WSID-100. The testing data can from the same categories manually labeled image dataset (\eg ImageNet).

\section{Conclusion}

In this work, we presented an automatic diverse image dataset construction framework. Our framework mainly involves three successive modules, namely multiple textual queries discovering, noisy textual queries filtering and noisy images filtering. Specifically, we first discover a set of semantically rich textual queries, from which the visual non-salient and less relevant textual queries are removed. To suppress the search error and noisy textual queries induced noisy images, we further divide the retrieved image noise into three types and use different methods to filter these noise separately. To verify the effectiveness of the proposed framework, we built an image dataset with 100 categories. 
Extensive experiments on the tasks of image classification and cross-dataset generalization have shown the superiority of our dataset over manually labeled datasets and web-supervised datasets.
In addition, we successfully applied our data to improve the object detection performance on the VOC 2007 dataset. The experimental results showed the superiority of our proposed work to several web-supervised and weakly supervised state-of-the-art methods. We have publicly released our web-supervised diverse image dataset on the website to facilitate the research in the web-vision and other related fields.

\ifCLASSOPTIONcaptionsoff
  \newpage
\fi

\vspace{-1.5cm}

\begin{IEEEbiography}
	[{\includegraphics[height=1.25in,clip,keepaspectratio]{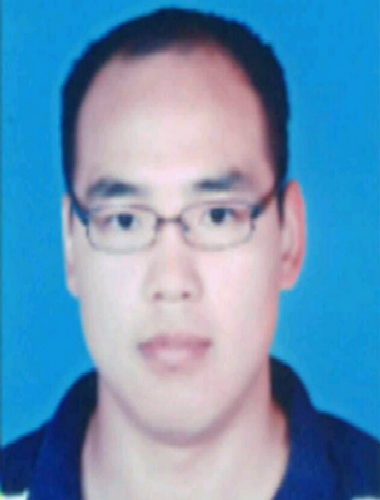}}]
	{Yazhou Yao}
	is a  research scientist in the Inception Institute of Artificial Intelligence (IIAI), Abu Dhabi, UAE. With the support of the China Scholarship Council, he received the PhD degree in Computer Science, University of Technology Sydney, Australia at 2018. Before that, he received the B.Sc. Degree and M.Sc. degree from Nanjing Normal University, China at 2010 and 2013 respectively. His research interests include social multimedia processing and machine learning.
\end{IEEEbiography}

\vspace{-1.5cm}

\begin{IEEEbiography}
	[{\includegraphics[height=1.25in,clip,keepaspectratio]{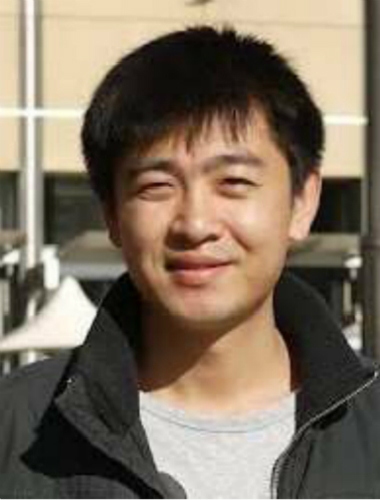}}]
	{Fumin Shen}
	received his Bachelor degree at 2007 and PhD degree at 2014 from Shandong University and Nanjing University of Science and Technology, China, respectively. Now he is a professor of Center for Future Media and School of Computer Science and Engineering, University of Electronic Science and Technology of China. His major research interests include computer vision and machine learning, including face recognition, image analysis and hashing methods.
\end{IEEEbiography}

\vspace{-1.5cm}

\begin{IEEEbiography}
	[{\includegraphics[height=1.25in,clip,keepaspectratio]{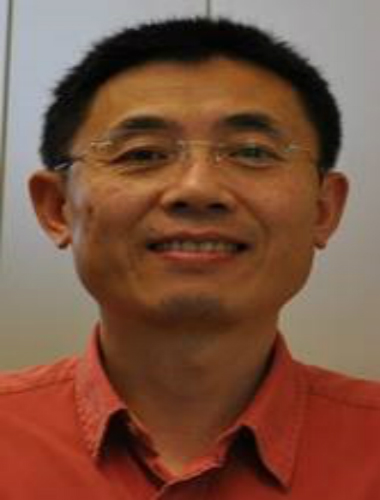}}]
	{Jian Zhang}
	(SM’04) received the B.{sc}. degree from East China Normal University, Shanghai, China, in 1982; the M.{sc}. degree in computer science from Flinders University, Adelaide, Australia, in 1994; and the Ph.D. degree in electrical engineering from the University of New South Wales (UNSW), Sydney, Australia, in 1999. He is currently an Associate Professor with the Faculty of Engineering and Information Technology, University of Technology Sydney, Sydney. His current research interests include Multimedia processing and communications, image and video processing, machine learning, pattern recognition, media and social media visual information retrieval and mining, human-computer interaction and intelligent video surveillance systems. He is Associated Editors for the IEEE Transactions on Multimedia.
\end{IEEEbiography}

\vspace{-1.5cm}

\begin{IEEEbiography}
	[{\includegraphics[width=1in,height=1.25in,clip,keepaspectratio]{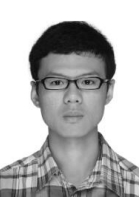}}]
	{Li Liu}
	received the B.Eng. degree in electronic information engineering from Xian Jiaotong University, Xian, China, in 2011, and the Ph.D. degree from the Department of Electronic and Electrical Engineering, University of Sheffield, Sheffield, U.K., in 2014. He is currently with the Inception Institute of Artificial Intelligence, Abu Dhabi, UAE. His current research interests include computer vision, machine learning, and data mining.
\end{IEEEbiography}

\vspace{-1.5cm}

\begin{IEEEbiography}
	[{\includegraphics[width=1in,height=1.25in,clip,keepaspectratio]{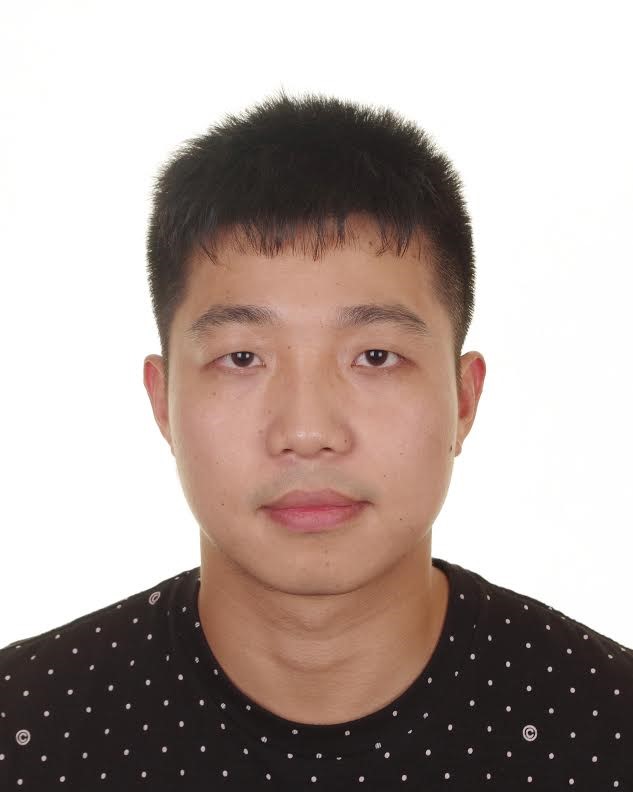}}]
	{Fan Zhu}
	received his MSc degree in Electrical Engineering with distinction, Ph.D degree in computer vision from the University of Sheffield, UK, in 2011 and 2015 respectively. He was a post-doctoral research fellow with the Electrical and Computer Engineering department of New York University Abu Dhabi and a data scientist with Pegasus LLC. He is now a Lead Scientist with the Inception Institute of Artificial Intelligence. His research interests include deep feature learning for 2D images and 3D shapes, scene understanding, video analytics and adversarial learning.
\end{IEEEbiography}

\vspace{-1.5cm}

\begin{IEEEbiography}
	[{\includegraphics[height=1.25in,clip,keepaspectratio]{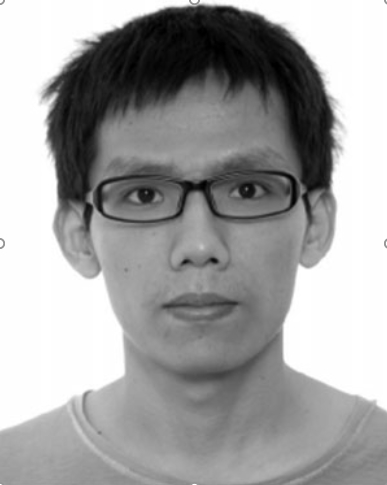}}]
	{Dongxiang Zhang}
	received the BSc degree from Fudan University, China, in 2006 and the PhD degree from the National University of Singapore, in 2012. He is a professor in the School of Computer Science and Engineering, University of Electronic Science and Technology of China. He worked as a research fellow with NExT Research
	Center, Singapore, from 2012 to 2014 and was promoted as a senior research fellow in 2015. His research interests include spatial databases, cloud computing, and big data analytics.
\end{IEEEbiography}

\vspace{-1.5cm}

\begin{IEEEbiography}
	[{\includegraphics[height=1.2in,keepaspectratio]{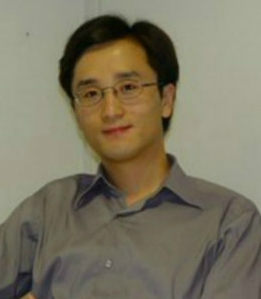}}]
	{Heng Tao Shen} 
	is a Professor of National ``Thousand Talents Plan'' and the director of Center for Future Media at the University of Electronic Science and Technology of China (UESTC). He obtained his BSc with 1st class Honours and PhD from Department of Computer Science, National University of Singapore (NUS) in 2000 and 2004 respectively. He then joined the University of Queensland (UQ) as a Lecturer, Senior Lecturer, Reader, and became a Professor in late 2011. His research interests mainly include Multimedia Search, Computer Vision, and Big Data Management on spatial, temporal, and multimedia databases. He has published over 150 peer-reviewed papers, most of which are in prestigious international venues of interests. For his outstanding research contributions, he received the Chris Wallace Award in 2010 conferred by Computing Research and Education Association, Australia, and the Future Fellowship from Australia Research Council in 2012. He is an Associate Editor of IEEE Transactions on Knowledge and Data Engineering, and has organized ICDE 2013 as Local Organization Co-Chair, and ACM Multimedia 2015 as Program Committee Co-Chair. 
\end{IEEEbiography}


\begin{thebibliography}{1}
	
	\bibitem{hua2015prajna}
	X.~Hua and J.~Li.
	\newblock ``Prajna: Towards recognizing whatever you want from images without image labeling,''
	\newblock {\em AAAI International Conference on Artificial Intelligence}, 137--144, 2015.
	
	\bibitem{schroff2011harvesting}
	F.~Schroff, A.~Criminisi, and A.~Zisserman.
	\newblock ``Harvesting image databases from the web,''
	\newblock {\em IEEE Transactions On Pattern Analysis and Machine Intelligence}, 33(4): 754--766, 2011.
	
	\bibitem{icme2016yao}
	Y.~Yao, J.~Zhang, F.~Shen, X.~Hua, J.~Xu, and Z.~Tang,
	\newblock ''Automatic image dataset construction with multiple textual metadata,''
	\newblock {\em IEEE International Conference on Multimedia and Expo}, 1--6, 2016.
	
	\bibitem{li2010optimol}
	L.~Li and L.~Fei-Fei,
	\newblock ``Optimol: automatic online picture collection via incremental model learning,''
	\newblock {\em International Journal of Computer Vision}, 88(2): 147--168, 2010.
	
	\bibitem{deng2009imagenet}
	J.~Deng, W.~Dong, R.~Socher, L.-J. Li, K.~Li, and L.~Fei-Fei,
	\newblock ``Imagenet: A large-scale hierarchical image database,''
	\newblock {\em IEEE International Conference on Computer Vision and Pattern Recognition}, 248--255, 2009.
	
	\bibitem{berg2006animals}
	Tamara~L Berg and David~A Forsyth,
	\newblock ``Animals on the web,''
	\newblock {\em IEEE International Conference on Computer Vision and Pattern Recognition}, 1463--1470, 2006.
	
	\bibitem{hare2010automatically}
	J.~Hare and P.~Lewis,
	\newblock ``Automatically annotating the mir flickr dataset: Experimental protocols, openly available data and semantic spaces,''
	\newblock {\em ACM International Conference on Multimedia Information Retrieval}, 547--556, 2010.
	
	\bibitem{fergus2004visual}
	Robert Fergus, Pietro Perona, and Andrew Zisserman,
	\newblock ``A visual category filter for google images,''
	\newblock {\em Europe Conference on Computer Vision}, 242--256, 2004.
	
	\bibitem{fergus2005learning}
	Robert Fergus, Li~Fei-Fei, Pietro Perona, and Andrew Zisserman,
	\newblock ``Learning object categories from google's image search,''
	\newblock {\em IEEE International Conference on Computer Vision}, 1816--1823, 2005.
	
	\bibitem{sindhwani2005co}
	V.~Sindhwani, P.~Niyogi, and M.~Belkin,
	\newblock ``A co-regularization approach to semi-supervised learning with
	multiple views,''
	\newblock {\em International Conference on Machine Learning}, 74--79, 2005.
	
	\bibitem{chang2011libsvm}
	C.-C. Chang and C.-J. Lin,
	\newblock ``Libsvm: a library for support vector machines,''
	\newblock {\em ACM Transactions on Intelligent Systems and Technology}, 2(3): 27, 2011.
	
	\bibitem{cilibrasi2007google}
	R.~Cilibrasi and P.~Vitanyi,
	\newblock ``The google similarity distance,''
	\newblock {\em IEEE Transactions on Knowledge and Data Engineering}, 19(3): 370--383, 2007.
	
	\bibitem{zhang2015memory}
	Hao Zhang, Gang Chen, Beng~Chin Ooi, Kian-Lee Tan, and Meihui Zhang,
	\newblock ``In-memory big data management and processing: A survey,''
	\newblock {\em IEEE Transactions on Knowledge and Data Engineering}, 27(7): 1920--1948, 2015.
	
	\bibitem{li2006sentence}
	Yuhua Li, David McLean, Zuhair~A Bandar, James~D O'shea, and Keeley Crockett,
	\newblock ``Sentence similarity based on semantic nets and corpus statistics,''
	\newblock {\em IEEE Transactions on Knowledge and Data Engineering}, 18(8): 1138--1150, 2006.
	
	\bibitem{divvala2014learning}
	S.~Divvala, C.~Guestrin,
	\newblock ``Learning everything about anything: Webly-supervised visual concept learning,''
	\newblock {\em IEEE International Conference on Computer Vision and Pattern Recognition}, 3270--3277, 2014.
	
	\bibitem{krizhevsky2012imagenet}
	Alex Krizhevsky, Ilya Sutskever, and Geoffrey~E Hinton,
	\newblock ``Imagenet classification with deep convolutional neural networks,''
	\newblock {\em Advances in Neural Information Processing Systems}, 1097--1105, 2012.
	
	\bibitem{chen2006miles}
	Y.~Chen, J.~Bi, and J.~Wang,
	\newblock ``Miles: Multiple-instance learning via embedded instance selection,''
	\newblock {\em IEEE Transactions on Pattern Analysis and Machine Intelligence}, 28(12): 1931--1947, 2006.
	
	\bibitem{everingham2010pascal}
	M.~Everingham, L.~Van~Gool, C.~K. Williams, J.~Winn, and A.~Zisserman,
	\newblock ``The pascal visual object classes (voc) challenge,''
	\newblock {\em International Journal of Computer Vision}, 88(2): 303--338, 2010.
	
	\bibitem{felzenszwalb2010object}
	P.~Felzenszwalb, R.~Girshick, and D.~Ramanan,
	\newblock ``Object detection with discriminatively trained part-based models,''
	\newblock {\em IEEE Transactions on Pattern Analysis and Machine Intelligence}, 32(9): 1627--1645, 2010.
	
	\bibitem{michel2011quantitative}
	J.-B. Michel, Y.~K. Shen, A.~P. Aiden, A.~Veres, M.~K. Gray, J.~P. Pickett,
	D.~Hoiberg, D.~Clancy, P.~Norvig, J.~Orwant, et~al.
	\newblock Quantitative analysis of culture using millions of digitized books.
	\newblock {\em Science}, 331(6014): 176--182, 2011
	
	\bibitem{krizhevsky2009learning}
	A.~Krizhevsky and G.~Hinton,
	\newblock ``Learning multiple layers of features from tiny images,''
	\newblock {\em Citeseer}, 2009.
	
	\bibitem{brefeld2006efficient}
	U.~Brefeld, T.~Scheffer, and S.~Wrobel,
	\newblock ``Efficient co-regularised least squares regression,''
	\newblock {\em ACM International Conference on Machine
		Learning}, 137--144, 2006.
	
	\bibitem{lin2012syntactic}
	Y.~Lin, J.~Michel, E.~Aiden, J.~Orwant, W.~Brockman, and S.~Petrov,
	\newblock ``Syntactic annotations for the google books ngram corpus,''
	\newblock {\em ACL 2012 System Demonstrations}, 169--174, 2012.
	
	\bibitem{miller1995wordnet}
	G.~A. Miller.
	\newblock ``Wordnet: a lexical database for english,''
	\newblock {\em Communications of the ACM}, 38(11): 39--41, 1995.
	
	\bibitem{siva2011weakly}
	P.~Siva and T.~Xiang,
	\newblock ``Weakly supervised object detector learning with model drift detection,``
	\newblock {\em IEEE International Conference on Computer Vision}, 343--350, 2011.
	
	\bibitem{torralba2011unbiased}
	A.~Torralba and A.~Efros.
	\newblock ``Unbiased look at dataset bias,``
	\newblock {\em IEEE International Conference on Computer Vision and Pattern Recognition}, 1521--152, 2011.
	
	\bibitem{vijayanarasimhan2008keywords}
	S.~Vijayanarasimhan and K.~Grauman.
	\newblock ``Keywords to visual categories: Multiple-instance learning for weakly supervised object categorization,``
	\newblock {\em IEEE International Conference on Computer Vision and Pattern Recognition}, 1--8, 2008.
	
	\bibitem{coates2011analysis}
	A.~Coates, A.~Ng, H.~Lee, 
	\newblock ''An analysis of single-layer networks in unsupervised feature learning,'' 
	\newblock {\em International Conference on Artificial Intelligence and Statistics}, 215--223, 2011.
	
	\bibitem{duan2011improving}
	L.~Duan, W.~Li, I.~Tsang, and D.~Xu,
	\newblock ``Improving web image search by bag-based reranking,''
	\newblock {\em IEEE Transactions on Image Processing}, 20(11): 3280--3290, 2011.
	
	\bibitem{griffin2007caltech}
	G.~Griffin, A.~Holub, P.~Perona, 
	\newblock ``Caltech-256 object category dataset.''
	
	\bibitem{prest2012learning}
	A.~Prest, C.~Leistner, J.~Civera, and V.~Ferrari,
	\newblock ``Learning object class detectors from weakly annotated video,''
	\newblock {\em IEEE International Conference on Computer Vision and Pattern Recognition}, 3282--3289, 2012.
	
	\bibitem{yao2017exploiting}
	Y.~Yao, J.~Zhang, F.~Shen, X.~Hua, J.~Xu, and Z.~Tang,
	\newblock ``Exploiting web images for dataset construction: A domain robust approach,''
	\newblock {\em IEEE Transactions on Multimedia}, 19(8): 1771--1784, 2017.
	
	\bibitem{renNIPS15fasterrcnn}
	S Ren, K He, R Girshick, and J Sun.
	\newblock Faster {R-CNN}: Towards Real-Time Object Detection with Region Proposal Networks
	\newblock {\em Advances in Neural Information Processing Systems ({NIPS})}, 91--99, 2015.	
	
	\bibitem{speer2013conceptnet}
	R.~Speer, C.~Havasi, 
	\newblock ``Conceptnet 5: A large semantic network for relational knowledge,''
	\newblock {\em The People’s Web Meets NLP}, 161--176, 2013.
	
	\bibitem{mm2016yao}
	Y.~Yao, X.~Hua, F.~Shen, J.~Zhang, and Z.~Tang,
	\newblock ``A domain robust approach for image dataset construction,``
	\newblock {\em ACM International Conference on Multimedia}, 212--216, 2016.
	
	\bibitem{Girshick2014}
	R Girshick, J Donahue, T Darrell, and J Malik.
	\newblock Rich Feature Hierarchies for Accurate Object Detection and Semantic Segmentation
	\newblock {\em IEEE Conference on Computer Vision and Pattern Recognition (CVPR)}, 580--587, 2014.
	
	\bibitem{hariharan2012discriminative}
	B Hariharan, J Malik, and D Ramanan,
	\newblock ``Discriminative decorrelation for clustering and classification,''
	\newblock {\em European Conference on Computer Vision}, 459--472, 2012.
	
	\bibitem{zitnick2014edge}
	C Zitnick, and  P Doll,
	\newblock ``Edge boxes: Locating object proposals from edges,''
	\newblock {\em European Conference on Computer Vision}, 391--405, 2014.	
	
	\bibitem{chen2014enriching}
	X Chen, A Shrivastava, and A Gupta,
	\newblock ``Enriching visual knowledge bases via object discovery and segmentation,''
	\newblock {\em IEEE Conference on Computer Vision and Pattern Recognition}, 2027--2034, 2014.
	
	\bibitem{ballan2015data}
	L Ballan, M Bertini, G Serra, A Delbimbo 
	\newblock ``A data-driven approach for tag refinement and localization in web videos,''
	\newblock {\em Computer Vision and Image Understanding}, 140: 58--67, 2015.
	
	\bibitem{ballan2012combining}
	L Ballan, M Bertini, A Delbimbo, A. M. Serain 
	\newblock ``Combining generative and discriminative models for classifying social images from 101 object categories,''
	\newblock {\em IEEE International Conference on Pattern Recognition}, 1731--1734, 2012.
	
	\bibitem{fei2006one}
	F Li, F Rob,  and P Pietro 
	\newblock ``One-shot learning of object categories,''
	\newblock {\em IEEE transactions on pattern analysis and machine intelligence}, 28(4): 594--611, 2006.
	
	\bibitem{tsikrika2012building}
	T Tsikrika, J Kludas, and A Popescu,
	\newblock ``Building reliable and reusable test collections for image retrieval: The wikipedia task at imageclef,''
	\newblock {\em IEEE MultiMedia}, 19(3): 24--33, 2012.
	
	\bibitem{mediaeval2017}
	M Zaharieva, B Ionescu, A.L. G{\^\i}nsca, R.L.T. Santos, and H M{\"u}ller,
	\newblock ``Retrieving Diverse Social Images at MediaEval 2017: Challenges, Dataset and Evaluation''
	\newblock {\em Proceedings of the MediaEval 2017 Workshop}, 2017.
	
	\bibitem{thomee2016yfcc100m}
	B Thomee, D.A. Shamma, G Friedland, B Elizalde, K Ni, D Poland, D Borth, and L.J. Li,
	\newblock ``YFCC100M: The new data in multimedia research''
	\newblock {\em Communications of the ACM}, 59(2): 64--73, 2016.
	
	\bibitem{goodfellow2014generative}
	I Goodfellow, J Pouget-Abadie, M Mirza, B Xu, D Warde-Farley, S Ozair, A Courville, Y Bengio
	\newblock ``Generative adversarial nets''
	\newblock {\em Advances in neural information processing systems}, 2672--2680, 2014.
	
	\bibitem{valcarce2016efficient}
	D Valcarce, J Parapar, A Barreiro
	\newblock ``Efficient Pseudo-Relevance Feedback Methods for Collaborative Filtering Recommendation''
	\newblock {\em European Conference on Information Retrieval}, 602--613, 2016.	

\end{thebibliography}
\end{document}